\documentclass[journal]{IEEEtran}

\def\baselinestretch{0.97}

\usepackage{cite}
\ifCLASSINFOpdf
\usepackage[pdftex]{graphicx}
\DeclareGraphicsExtensions{.pdf,.jpeg,.png}
\else
\usepackage[dvips]{graphicx}
\DeclareGraphicsExtensions{.eps}
\fi
\usepackage{epsfig,epsf,epstopdf,graphicx}
\usepackage[cmex10]{amsmath}
\usepackage{amssymb}
\usepackage{amsthm }
\usepackage{nicefrac}
\usepackage{hyperref}
\hypersetup{colorlinks=true}
\usepackage{tabularx}
\newtheorem{theorem}{Theorem}
\newtheorem{lemma}{Lemma}
\theoremstyle{theorem}
\newtheorem{Corollary}{Corollary}
\theoremstyle{definition}
\newtheorem{Definition}{Definition}
\theoremstyle{plain}
\newtheorem{Proposition}{Proposition}
\theoremstyle{plain}

\newcommand{\avec}{{\bf{a}}}
\newcommand{\bvec}{{\bf{b}}}
\newcommand{\cvec}{{\bf{c}}}
\newcommand{\dvec}{{\bf{d}}}
\newcommand{\evec}{{\bf{e}}}
\newcommand{\hvec}{{\bf{h}}}
\newcommand{\nvec}{{\bf{n}}}
\newcommand{\rvec}{{\bf{r}}}
\newcommand{\svec}{{\bf{s}}}
\newcommand{\uvec}{{\bf{u}}}
\newcommand{\vvec}{{\bf{v}}}
\newcommand{\xvec}{{\bf{x}}}
\newcommand{\yvec}{{\bf{y}}}
\newcommand{\zvec}{{\bf{z}}}
\newcommand{\zerovec}{{\bf{0}}}
\newcommand{\onevec}{{\bf{1}}}

\newcommand{\Real}{\mathbb{R}}
\newcommand{\Expv}{\mathbb{E}}
\newcommand{\var}{\mathrm{\mathbf{var}}}

\newcommand{\mub}{{\mbox{\boldmath $\mu$}}}

\newcommand{\Ab}{\mathbf{A}}
\newcommand{\Bb}{\mathbf{B}}
\newcommand{\Cb}{\mathbf{C}}
\newcommand{\Db}{\mathbf{D}}
\newcommand{\Gb}{\mathbf{G}}
\newcommand{\Hb}{\mathbf{M}}
\newcommand{\Ib}{\mathbf{I}}
\newcommand{\Kb}{\mathbf{K}}
\newcommand{\Mb}{\mathbf{M}}

\newcommand{\Pb}{\mathbf{P}}
\newcommand{\Qb}{\mathbf{Q}}
\newcommand{\Rb}{\mathbf{R}}
\newcommand{\Tb}{\mathbf{T}}
\newcommand{\Wb}{\mathbf{W}}

\newcommand{\Gaus}{{\cal{N}}}
\newcommand{\norm}{\vert \vert}
\newcommand{\lag}{{\cal L}}
\newcommand{\lzero}{$\ell_{0}$}
\newcommand{\lone}{$\ell_{1}$}
\newcommand{\ltwo}{$\ell_{2}$}
\newcommand{\CSIM}{{\mathrm{CSIM}}}

\makeatletter
\newcommand{\pushright}[1]{\ifmeasuring@#1\else\omit\hfill$\displaystyle#1$\fi}
\newcommand{\pushleft}[1]{\ifmeasuring@#1\else\omit$\displaystyle#1$\hfill\fi}
\makeatother

\usepackage{algorithm}
\usepackage{algorithmic}
\usepackage{multirow}

\usepackage[caption=false,font=footnotesize]{subfig}

\usepackage{fixltx2e}
\usepackage{afterpage}
\usepackage{float}
\usepackage{stfloats}
\usepackage[bottom]{footmisc}
\usepackage{url}
\begin{document}

\title{A Convex Similarity Index for Sparse Recovery of Missing Image Samples}

\author{Amirhossein~Javaheri, Hadi~Zayyani,~\IEEEmembership{Member,~IEEE}~and~Farokh~Marvasti,~\IEEEmembership{Life Senior Member,~IEEE}% <-this % stops a space
\thanks{Amirhossein~Javaheri and Farokh~Marvasti are both with Sharif~University~of~Technology, Tehran, Iran. (email:~javaheri\_amirhossein@ee.sharif.edu; email: marvasti@sharif.edu).}% <-this %
\thanks{H. Zayyani is with the Department
of Electrical and computer Engineering, Qom University of Technology (QUT), Qom, Iran (e-mail: zayyani@qut.ac.ir).}%<-this %
% <-this %
}

\maketitle

\begin{abstract}
This paper investigates the problem of recovering missing samples using methods based on sparse representation adapted especially for image signals. Instead of \ltwo-norm or Mean Square Error (MSE), a new perceptual quality measure is used as the similarity criterion between the original and the reconstructed images. The proposed criterion called Convex SIMilarity (CSIM) index is a modified version of the Structural SIMilarity (SSIM) index, which despite its predecessor, is convex and uni-modal. We derive mathematical properties for the proposed index and show how to optimally choose the parameters of the proposed criterion, investigating the Restricted Isometry (RIP) and error-sensitivity properties. We also propose an iterative sparse recovery method based on a constrained \lone-norm minimization problem, incorporating CSIM as the fidelity criterion. The resulting convex optimization problem is solved via an algorithm based on Alternating Direction Method of Multipliers (ADMM). Taking advantage of the convexity of the CSIM index, we also prove the convergence of the algorithm to the globally optimal solution of the proposed optimization  problem, starting from any arbitrary point. Simulation results confirm the performance of the new similarity index as well as the proposed algorithm for missing sample recovery of image patch signals.
\end{abstract}


\section{Introduction}
\label{sec:Introduction}

The missing sample recovery problem arises in many applications in the literature of signal processing \cite{Wang05,Stan14,Stoi09}. It is also known as inpainting in the context of audio and image processing. Audio inpainting is investigated in \cite{Adler12}, while image inpainting is discussed in \cite{Guler06} and \cite{Guil14}.

Among numerous algorithms for missing sample recovery and inpainting, some of them exploit the sparsity of the signals \cite{Stan14,Li14}. In this paper, we restrict ourselves to these sparse representation based algorithms. In this regard, it is assumed that the signal is sparse within a domain such as Discrete Cosine Transform (DCT), Discrete Wavelet Transform (DWT) \cite{WP95} or any other predefined or learned complete or overcomplete dictionary. The sparsity of the signal on a dictionary-based representation, means that the vector of coefficients of the signal in the transform domain has many zeros (or nearly zeros) and only a few of its elements are nonzero. Neglecting the insignificant (zero) coefficients, it is possible to reconstruct the signal with considerably low error. The sparsity of the signal gives us the ability to reconstruct it from very few random measurements far below the Nyquist rate. This is well known as Compressed Sensing (CS) \cite{Dono06,CandT06}, which has had many applications in the past decade \cite{Marv12}. The problem of reconstruction of the sparse signal from a few random measurements, is also known as sparse recovery. Many algorithms are proposed for sparse recovery of signals in different applications in audio and image processing.

In a missing sample recovery problem, some samples of the signal are missed due to physical impairment, unavailability of measurements, or distortion and disturbances. In such cases, it is shown that the corrupted samples would better be omitted throughout the reconstruction process \cite{Stan14}. Thus, the discarded samples may be considered as missed. Even with these missing samples, the signal can still be reconstructed, given the sparsifying basis or dictionary and the corresponding sparse coefficients. Many algorithms and optimization problems are suggested to recover the sparse samples in this regard. The fundamental problem in a sparse recovery method is to maximize the sparsity which is principally stated in terms of the \lzero-norm. There are a class of greedy algorithms for strictly sparse signal recovery based on \lzero-norm minimization. These include Matching Pursuit (MP) \cite{MallZ93}, Orthogonal Matching Pursuit (OMP) \cite{Trop07}, Regularized OMP (ROMP) \cite{Need09}, Compressive Sampling Matching Pursuit (CoSaMP) \cite{Need09COSAMP} and Generalized Matching Pursuit (GOMP) \cite{GOMPWang}. There are also iterative methods based on majorization minimization technique proposed for approximate \lzero-norm minimization using surrogate functions. The Iterative Hard Thresholding (IHT) algorithm \cite{IHT_Surr} is the first member of this class. There is also a modified version which uses adaptive thresholding named as Iterative Method with Adaptive Thresholding (IMAT) \cite{Marv003} with different variants including IMATI and IMATCS \cite{Azgh13}. A recent improved version of this method called INPMAT is also proposed in \cite{INPMAT}. Furthermore there is an approach for sparse approximation based on Smoothed-\lzero~(SL0) norm minimization presented in \cite{Mohi09}. The \lzero-minimization algorithms are mostly used in cases where the signal has exactly sparse support and the sparsity is known. But in many practical situations the sparsity is unknown or the signal is not strictly sparse but instead compressible, meaning that most coefficients are negligible (despite being precisely zero) compared to the significant elements. A good and a common alternative is to use the \lone-norm as the nearest convex approximation of \lzero-norm. This approach is called \lone-minimization or the basis-pursuit method \cite{ChenDS99}. There are many algorithms presented for \lone-norm minimization including Iterative Soft Thresholding Algorithm (ISTA) \cite{Daub04} and the fast version FISTA \cite{Beck09}, \lone~Least Squares (L1-LS) \cite{BoydTrunc}, Primal and Dual Augmented Lagrangian Methods (PALM and DALM)\cite{Yang11}, Iterative Bayesian Algorithm (IBA) \cite{ZayyBJ09}, Sparse Bayesian Learning (SBL) \cite{SBL} and Bayesian Compressed Sensing (BCS) \cite{BCS}. There are also more general $p$-norm minimization based algorithms available for solving the sparse recovery problem in the literature \cite{Char07}. For detailed survey on sparse recovery methods, one can refer to \cite{Zhang15}.

In this paper, we propose an alternative \lone-minimization method for sparse recovery of signals. In particular, we consider the sparse recovery of image patches with missing samples which has application in image inpainting and restoration. We use a criterion for measuring the similarity between two image signals, called Convex SIMilarity (CSIM) Index \cite{SAMPTA_Jav}. Although it is derived from the Structural SIMilarity (SSIM) index, the well-known perceptual quality assessment criterion \cite{SSIM04}, it has feasible mathematical features unlike its predecessor. These properties including convexity and error-sensitivity, which are investigated in this paper, result in simplified methods for solving the optimization problem incorporating CSIM index as the measure of similarity. We also present analyses on how to choose optimal parameters of the similarity index proposed in \cite{SAMPTA_Jav}. In this paper, we use CSIM criterion in our proposed optimization problem for missing sample recovery. Similar to \cite{Yang11}, an iterative algorithm is presented for \lone-minimization based on Alternating Direction Method of Multipliers (ADMM) method. Simulation results show the efficiency of the proposed method called CSIM minimization via Augmented Lagrangian Method (CSIM-ALM) compared to some popular existing algorithms.

\section{Metrics for Image Quality Assessment}
\label{sec:QualityAssessment}
%\subsection{A brief literature survey}
There are different criteria for Image Quality Assessment (IQA) which can be used as measures of similarity between two signals. The most popular fidelity metric for measuring the similarity between two signals $\xvec, \yvec\in \Real^n$ is MSE which is defined as $\mathrm{MSE}(\xvec,\yvec)=\frac{1}{n}\norm \xvec-\yvec\norm^2_2$. This criterion is widely used, because MSE or equivalently \ltwo-norm is mathematically a well-defined function of the difference between the reference and the test signal. This function has desirable mathematical features such as convexity and differentiability. Hence,  the optimization problem is simple and tractable using convex programming methods.
Nevertheless, there are cases in which the MSE criterion seems to be inefficient to accurately recover the original image signal, especially in the presence of noise. One reason is that this fidelity metric is indifferent toward the error distribution, i.e., the statistics of the error signal. For instance consider two scenarios in which a signal is corrupted. In the first case the reference signal is perturbed by noise and in the second, a constant amplitude is added to the original image. Both corrupted images have the same MSE distance with respect to the original signal, whilst the noisy image is definitely more visually deteriorated. Thus, there are a class of perceptual criteria introduced for measuring visual quality of images. The most popular perceptual metric is SSIM which is defined as \cite{SSIM04}:
\begin{equation}
\label{eq_SSIM}
\mathrm{SSIM}(\xvec,\yvec) =\Big( \dfrac{2\mu_x \mu_y+C_1}{\mu_x^2+\mu_y^2+C_1} \Big) \Big(
\dfrac{2\sigma_{x,y}+C_2}{\sigma_x^2+\sigma_y^2+C_2} \Big)
\end{equation}
where $C_1, C_2>0$ are constant and $\mu$ and $\sigma$ denote the mean and the variance (cross-covariance), respectively. This function whose mathematical properties are discussed in \cite{SSIM12}, is non-convex and multi-modal implying that the problem of optimizing this criterion is hard to solve.

\subsection{The CSIM index}
\label{Sec:ProposedMetric}
As mentioned earlier, SSIM function is non-convex and multi-modal which results in hard optimization problems. In \cite{SAMPTA_Jav}, we have proposed a simplified criterion which inherits error-sensitivity property of SSIM.
The proposed index called CSIM, is defined as follows:
\begin{equation}
\label{eq_1}
\mathrm{CSIM}(\xvec, \yvec) = k_1 (\mu_x^2 + \mu_y^2 - 2\mu_x \mu_y)+k_2 (\sigma_x^2+\sigma_y^2-2\sigma_{x,y})
\end{equation}
where $k_1$ and $k_2$ are positive constants and $\mu$ and $\sigma$ denote statistical estimates for mean and variance/cross-covariance. It is assumed that $k_1<k_2$, to ensure biased sensitivity toward random disturbance or noise compared to uniform level change. In fact unlike MSE, the proposed criterion has noise-sensitive variation. In other words, a constant change in the brightness level of the image does not alter this criterion as much as noise. This is logically because constant change in the amplitude only affects the signal mean value; it does not change variance/cross-covariance. Thus, as far as $k_1< k_2$, the CSIM index is only slightly influenced by mere brightness level change.
This index also benefits some feasible mathematical features, including convexity and differentiability.

\begin{theorem}
\label{theo:2}
For positive $k_1$ and $k_2$, the fidelity criterion defined by equation \eqref{eq_1} is strictly convex with respect to $\xvec$ or $\yvec$. Furthermore, there exists a transformation matrix $\Tb$, such that the proposed criterion can be restated in terms of \ltwo-norm, after the transformation.
\end{theorem}
\begin{IEEEproof}
Suppose $\xvec, \yvec \in \mathbb{R}^n$. If we use unbiased estimate for variance and covariance, we will have:
\begin{equation}
\label{eq_2}
\sigma_x^2 = \frac{1}{n-1} \norm \xvec-\mub_x \norm_2^2, \quad \sigma_{x,y} =\frac{1}{n-1}(\xvec-\mub_x)^T (\yvec-\mub_y)
\end{equation}
where $\mub_x=\mu_x \onevec_n$ and $\onevec_n = (1, \ldots, 1)^T \in \Real ^n$. Therefore, $\sigma_x^2 +\sigma_y^2-2\sigma_{x,y} = \frac{1}{n-1} \norm (\xvec-\mub_x)-(\yvec-\mub_y) \norm_2^2
$ and $\mu_x^2+\mu_y^2-2\mu_x \mu_y$ equals $(\mu_x-\mu_y)^2$. Hence, if we define the error signal as the difference between test signals, i.e., $\evec=\xvec-\yvec$, equation \eqref{eq_1} is simplified to:
\begin{align}
\label{eq_5}
\CSIM(\xvec,\yvec)&=\CSIM(\evec)=k_1 \mu_e^2+ \frac{k_2}{n-1} \norm \evec-\mu_e \onevec_n \norm_2^2
\end{align}
which implies $\CSIM$ is always non-negative.
Likewise, if we consider $\xvec$ and $\yvec$ as samples of random variables $X$ and $Y$, we have  $\CSIM(X,Y)=k_1 \Expv^2[X-Y] + k_2 \var[X-Y] \geq 0$ which equals zero if and only if $P(X=Y)=1$.
Now, using simplifications in \eqref{eq_2}, one can expand equation \eqref{eq_5} as follows:
\begin{align}
\label{eq_6}
%\mu_e &= \frac{1}{N} \onevec_N^T \evec = \frac{1}{N} \evec^T \onevec_N \\
\CSIM(\evec) &= \frac{k_1}{n^2} \evec^T \onevec_n \onevec_n^T \evec +\frac{k_2}{n-1}\evec^T \Mb^T \Mb \evec \nonumber \\
&= \evec^T\!\!\left(\frac{k_1}{n^2} \onevec_n \onevec_n^T+\frac{k_2}{n-1}\Mb\right)\!\evec=\evec^T \Wb \evec
\end{align}
where $\Mb= \Mb^T \Mb=\Ib_n- \frac{1}{n}\onevec_n \onevec_n^T$ and $\Ib_n$ is the identity matrix of size $n$.
%Therefore, we have:
%\begin{equation}
%\label{eq_7}
%\CSIM(\xvec,\yvec) = \CSIM(\evec)=\evec^T \Wb \evec
%\end{equation}
Therefore, the matrix $\Wb \in \mathbb{R}^{n\times n}$ is obtained by:
\begin{equation}
\label{eq_8}
\Wb = \frac{k_2}{n-1}\Ib_n+\left( \frac{k_1}{n^2}-\dfrac{k_2}{n(n-1)}\right)  \onevec_n \onevec_n^T
 = \theta_1 \Ib_n + \theta_2 \onevec_n \onevec_n^T
\end{equation}
Since $\CSIM$ is a quadratic function of $\evec$, the positive-definiteness of the Hessian matrix $\Wb$, implies its strict convexity as well. On the other hand, since $\evec=\xvec-\yvec$ is affine, it will also be concluded that $\CSIM$ is strictly convex with respect to both $\xvec$ and $\yvec$.
%According to \eqref{eq_6}, the Hessian is equal to $\Wb$. Now, using the proof in Appendix \ref{section:App_eigen}, the sorted eigenvalues of $\Wb$ are calculated:
%\begin{equation}
%\label{eq_13}
%\lambda_{\Wb_i}=\left\{\!\!\begin{array}{ll}
%\frac{k_2}{n-1},& i< n\\
%\frac{k_1}{n},& i=n
%\end{array}\right.
%\end{equation}
%As far as $k_1,\, k_2>0$, it is implied that $\Wb$ is positive-definite and consequently $\CSIM$ is strictly convex.
Now, to prove the second part of the theorem, we proceed as follows. Since $\Wb$ is Hermitian and positive-definite, it is diagonalizable and thus, $\Wb^{1/2}$ exists:
\begin{equation}
\label{eq_W_half}
\Wb^{1/2}=\sqrt{\frac{k_2}{n-1}} \Ib+ \frac{1}{n}\Bigg(\sqrt{ \frac{k_1}{n}}-\sqrt{\frac{k_2}{n-1} }\Bigg)\onevec_n\onevec_n^T
\end{equation}
%where $\beta_1$ and $\beta_2$ are obtained by:
%\begin{align}
%\beta_1 & = \sqrt{\frac{\rho}{n-1}}  \\
%\beta_2 & = \frac{1}{n}\Big(\sqrt{ \frac{1}{n}}-\sqrt{\frac{\rho}{n-1} }\Big)\nonumber
%\end{align}
Hence, we can write:
\begin{equation}
\label{eq_kernel}
\CSIM(\xvec,\yvec) =(\xvec-\yvec)^T {\Wb^{1/2}}^T {\Wb^{1/2}}(\xvec-\yvec) = \norm \Tb  (\xvec-\yvec) \norm_2
\end{equation}
which means there exists a transformation denoted by $\Tb=\Wb^{1/2}$, after which, we can restate the proposed index in terms of \ltwo-norm distance.
\end{IEEEproof}

\subsection{The choice of the parameters $k_1$ and $k_2$}
Now, the problem is how to find $k_1$ and $k_2$ so that the proposed criterion has utmost sensitivity toward random perturbation compared to uniform change.
\begin{Definition}
Assume $\evec_1\in \Real^n$ is a random binary signal with i.i.d. elements taking values $\{a,-a\}$ with equal probability, i.e., $C_{\evec_1} =\Expv[\evec_1\evec_1^T] =a^2 \Ib_n$, and $\evec_2=a \onevec_n$ is a deterministic constant-amplitude signal. Let $\yvec_1=\xvec+\evec_1$ and $\yvec_2 = \xvec+\evec_2$ where $\xvec \in \Real ^n$ is the reference signal.
We define the ratio of sensitivity as:
\label{Definition:1}
\begin{equation}
\label{eq_15}
r_S =\dfrac{\Expv[\CSIM(\xvec,\yvec_1)]}{\Expv[\CSIM(\xvec,\yvec_2)]} =\dfrac{\Expv[\CSIM(\evec_1)]}{\Expv[\CSIM(\evec_2)]} = \dfrac{\Expv[\evec_1^T \Wb \evec_1]}{\evec_2^T \Wb \evec_2}
\end{equation}
\end{Definition}
\begin{Proposition}
\label{Prop:2}
The sensitivity ratio with respect to CSIM criterion with kernel $\Wb$, is obtained by $r_S = \dfrac{\sum_i w_{i,i}}{\sum_i \sum_j w_{i,j}}$.
\end{Proposition}
\begin{IEEEproof}
Since $\evec_1^T \Wb \evec_1$ is scalar and $\Expv$ and $\mathrm{Trace}$ are linear operators, we may write:
\begin{align}
\label{eq_16}
\Expv[\evec_1^T \Wb \evec_1] &=\Expv[\mathrm{Trace}(\evec_1^T \Wb \evec_1)] \nonumber \\
&=\mathrm{Trace}(\Expv[\evec_1 \evec_1^T] \Wb) =a^2 \mathrm{Trace}(\Wb)
\end{align}
Thus:
\begin{align}
\label{eq_17}
\dfrac{\Expv[\evec_1^T \Wb \evec_1]}{\evec_2^T \Wb \evec_2} &=\dfrac{a^2 \mathrm{Trace}(\Wb)}{a^2 \onevec_n^T \Wb \onevec_n } = \dfrac{\sum_i w_{i,i}}{\sum_i \sum_j w_{i,j}}
\end{align}
\end{IEEEproof}
Now for $\Wb$ defined in \eqref{eq_9}, the sensitivity ratio is $r_S = \frac{k_2}{k_1}+\frac{1}{n}$.
This states that the greater the ratio $k_2/k_1$ is, the more sensitive to noise, the proposed CSIM index will be. But there are other conditions which impose constraints on the eigenvalues of $\Wb$ and accordingly the values of $k_1$ and $k_2$.
\subsubsection{Condition number}
Consider the optimization problem below with $\Db \in \Real ^{n\times p}$:
\begin{align}
\label{eq_18_2}
\min_{\svec} \CSIM(\Db \svec,\yvec)&= (\Db \svec-\yvec)^T\Wb(\Db \svec-\yvec) \\
&=\norm \Wb^{1/2}(\Db \svec-\yvec) \norm_2^2= \norm \Db' \svec-\yvec' \norm_2^2\nonumber
\end{align}
%Using \eqref{eq_kernel}, we can rewrite \eqref{eq_18_1} as:
%\begin{equation}
%
%\min_{\svec} \, (\Db \svec-\yvec)^T\Wb^{1/2}\Wb^{1/2}(\Db \svec-\yvec) = \norm \Db' \svec-\yvec' \norm_2^2
%\end{equation}
where $\Db' = \Wb^{1/2}\Db$ and $\yvec'=\Wb^{1/2}\yvec$. If $\Db'$ is full-column rank ($n>p$), the solution to \eqref{eq_18_2} will be obtained by
%$\Db'^{\dagger} \yvec' = \Db'^T (\Db'\Db'^T)^{-1}\yvec' = \Db^T \Wb^{1/2} (\Wb^{1/2}\Db\Db^T\Wb^{1/2})^{-1}  \Wb^{1/2}\yvec=\Db^{\dagger}  \yvec$.
$\Db'^{\dagger} \yvec' =  (\Db^T\Wb\Db)^{-1} \Db^T \Wb \yvec = \tilde{\Db} \yvec$\footnote{In case where $\Db'$ is full-row rank ($n<p$), we have $\Db'^{\dagger} \yvec'= \Db^{\dagger}  \yvec$. Thus, the matrix $\Wb$ does not change the condition number}.
To have a robust (reliable) solution, the matrix $\tilde{\Db}$ must not be ill-conditioned. Now assuming $\Db$ or equivalently $\tilde{\Db}$ is full-rank, using the extended definition of condition number for non-square matrices, we can write:
\begin{equation}
\label{eq_kappa_def}
\kappa(\tilde{\Db}) = \frac{\sigma_{\max}(\tilde{\Db})}{\sigma_{\min}(\tilde{\Db})}
\end{equation}
where $\sigma_{\min}(\tilde{\Db})$ denotes the minimum (non-zero) singular value of $\tilde{\Db}$. The following theorem states how the constraint on the maximum value of $\kappa(\tilde{\Db})$ corresponds to the ratio of the parameters $k_1$ and $k_2$.
\begin{theorem}
\label{theo:kappa}
Define $\xi=\kappa(\Db)\left( \frac{n}{n-1}\right) \left(\frac{1}{\kappa^2(\Db)}-\frac{1}{n} \frac{\onevec^T \Db \Db^T \onevec}{ \sigma^2_{\max}(\Db)}\right)$ and $\nu=\kappa(\Db)\left(\frac{1}{n} \frac{\onevec^T \Db \Db^T \onevec}{ \sigma^2_{\max}(\Db)}\right)$. The matrix $\tilde{\Db}$ satisfies $\kappa(\tilde{\Db})\leq\kappa_{\max}$ with $\kappa_{\max}>\xi+\nu$, if:
\begin{equation}
1<\frac{k_2}{k_1} \leq \frac{\kappa_{\max}-\nu}{\xi}
\end{equation}
\end{theorem}
\begin{IEEEproof}
Refer to Appendix \ref{section:App_kappa}.
\end{IEEEproof}

\subsubsection{RIP condition}
Another constraint is imposed when using the proposed criterion in a \lone-minimization problem. In particular, consider the Basis Pursuit Denoising (BPDN) problem \cite{BPDN} with CSIM as the fidelity criterion:
\begin{equation}
\label{eq_18_4}
\min_{\svec} \norm \svec \norm_1, \quad \text{s.t.}\quad \norm \Db' \svec-\yvec' \norm_2^2\leq \epsilon
\end{equation}
In order for \eqref{eq_18_4} to uniquely recover $k$-sparse signals, the matrix $\Db'$ should satisfy RIP of order $2k$ with constant $\delta_{2k}< \sqrt{2}-1$. Assume the columns of $\Db$ have unit norm and let $\mu(\Db)$ denote the mutual coherence between these column vectors. Investigating the RIP condition for $\Db'$, yields the following theorem:
\begin{theorem}
\label{theo:3}
Assume the columns of $\Db$ are normalized. The matrix $\Db'=\Wb^{1/2} \Db \in \Real ^{n\times p}$ satisfies RIP condition of order $2k$ with constant $\delta'_{2k}<\sqrt{2}-1$, if the following conditions are satisfied:
\begin{align}
1& <\frac{k_2}{k_1} \leq \frac{C_1}{C_2-\delta'_{2k}} \nonumber \\
n&> n_{\max} \nonumber \\
2<2k &\leq \min \left\{ \left(1+\frac{\delta'_{2k}}{\mu(\Db)}\right), \min \{n,p\} \right\}\end{align}
where
\begin{align}
\label{eq_RIP}
C_1 &= \frac{1}{n^2}(2k-1)(n-1)(n-1+\mu(\Db)) \nonumber \\
C_2 &= \frac{2k-1}{n}(n-1+\mu(\Db)(n+1)) \nonumber \\
n_{\max}&=\dfrac{1+\sqrt{1-4\left(\frac{\delta'_{2k}}{2k-1}-\mu(\Db)\right)\big(1-\mu(\Db)\big)}}{2\left(\frac{\delta'_{2k}}{2k-1}-\mu(\Db)\right)} \nonumber
\end{align}.
\end{theorem}
\begin{IEEEproof}
Refer to Appendix \ref{section:App_RIP} for the proof.
\end{IEEEproof}
\begin{Corollary}
\label{Corr1}
To gain the maximum ratio of sensitivity, while satisfying the condition number and the RIP restrictions, we choose:
\begin{equation}
\frac{k_2}{k_1} = \min\left\{\frac{C_1}{C_2-\delta'_{2k}}, \frac{\kappa_{\max}-\nu}{\xi}\right\}
\end{equation}
\end{Corollary}
where in practice, we choose $\kappa_{\max}=4$ and $\delta'_{2k}=0.4$. Furthermore, in case where the sparsity of the signal is unknown, we assume 10\% sparsity, i.e., $k=\lfloor0.1 n\rfloor$. For more details go to section \ref{sec:Simulations}.
%Now if we choose $\mu(\Db)=0$ in the numerator expression $C_1$ and $\mu(\Db)=\frac{1}{2k-1}$ in the denominator term $C_2-\delta'_{2k}$,
\section{Statement of The Problem}
\label{sec:Problem}
The problem of recovering an image with samples missed at random, is equivalent to random sampling reconstruction of a signal. This problem is also addressed in the literature as block loss restoration due to error in the transmission channel \cite{Talebi00,Hosseini14}. Or alternatively, known as image inpainting in applications where the sampling mask is known and the objective is to fill in the gaps or remove occlusion or specific objects from the image \cite{Bertalmio00, Guil14,Criminisi04}. Suppose $\xvec \in \mathbb{R}^N$ is the vectorized image signal and $\Hb$ is the sampling matrix by which the pattern of sampling of the image signal is determined.
%In other words $\Hb$ is a diagonal matrix with $0$s or $1$s (respectively corresponding to missed and observed samples) on its main diagonal. The observed image signal with missing samples, is also denoted by $\yvec \in \Real^N$.
In other words $\Hb\in \Real^{N\times N}$ is the identity matrix with $N-m$ of its diagonals set to 0, where $m$ denotes the number of available samples. The observed image signal with missing samples, is also denoted by $\yvec \in \Real^N$.
Among many approaches for missing sample recovery of images, there are also a class of inpainting algorithms which use sparse representation for image restoration. In \cite{Guler06, Azgh13, Elad2005, Fadili07}, there are iterative algorithms proposed for recovery of missing samples exploiting the sparsity of representation based on redundant (over-complete) dictionaries. If we assume that $\xvec$ has approximately a sparse representation based on the atoms of a dictionary specified by the matrix $\Db$, the regular optimization problem for sparse recovery of the missing samples is formulated as follows:
\begin{equation}
\label{eq_19}
\min_{\xvec,\svec} \norm \svec \norm_1 \quad \text{s.t.}\quad \left\{\begin{array}{l}
\norm \Hb\xvec-\yvec \norm_2 \leq \epsilon_n \\
\xvec=\Db \svec%
\end{array}\right.
\end{equation}
where $\svec$ denotes the sparse vector of representation coefficients and $\epsilon_n$ denotes the variance of the additive observation noise. This problem is indeed the extension of the BPDN problem for missing sample recovery.
Using a proper value of $\alpha$, this problem is also shown to be equivalent to \cite{BPDN}:
\begin{equation}
\label{eq_19_1}
\min_{\xvec,\svec} \norm \Hb\xvec-\yvec \norm_2^2+ \alpha \norm \svec \norm_1 \quad \text{s.t.}\quad\xvec=\Db \svec
\end{equation}

Also, there exist methods optimized in terms of different image quality measures. In \cite{Oga13}, an exemplar-based method for image completion is proposed which uses dictionary learning based on SSIM for spare recovery of local image patches. In this algorithm, for each candidate patch, the optimization problem below is solved within the sparse coding step:
\begin{equation}
\label{eq_18_9}
\max_{\xvec,\svec}
\mathrm{SSIM}(\xvec, \Db \svec) \quad \text{s.t.}\quad \left\{\!\!\begin{array}{ll}
 \Hb \xvec=\yvec \\
\norm \svec\norm_0 \leq T
\end{array}\right.
\end{equation}
This problem is iteratively solved using a matching pursuit approach, i.e., in each step, the support of the sparse vector is retrieved and the coefficients are subsequently obtained solving unconstrained \eqref{eq_18_9}. Although this problem addresses perceptual quality of reconstruction, it is non-convex and thus solved using time-consuming linear search methods.
In this paper, we propose an alternative method for sparse recovery of image signals which can be applied in the sparse coding step of an adaptive dictionary learning method for image inpainting. We also use a perceptual metric, instead of \ltwo-norm, for visually enhanced reconstruction of the missing samples.

\section{The Proposed Algorithm}
\label{sec:SparseRecovery}
As discussed earlier in section \ref{sec:Problem}, most of the algorithms use \ltwo-norm as fidelity criterion for image reconstruction. But, there are also inpainting methods based on local sparse representation which use perceptual IQA metrics namely SSIM, for recovery of the missing samples. Here, we propose to use CSIM instead of SSIM.
Hence, to solve the missing sample recovery problem defined in (\ref{eq_19}), we incorporate our proposed perceptual metric for reconstruction of the image samples. In fact, among all the solutions with equal \ltwo-norm distance from $\yvec$, we seek for those with maximum similarity in terms of $\CSIM$. Hence, the proposed optimization problem is as follows:
\begin{equation}
\label{eq_20}
\min_{\xvec,\svec}
\CSIM(\Hb\Db \svec,\yvec) \quad \text{s.t.}\quad
\left\{\!\!\begin{array}{ll}
\norm \Hb \Db \svec-\yvec\norm<\epsilon_n\\
\norm \svec \norm_1  \leq T
\end{array}\right.
\end{equation}
which is equivalent to
\begin{equation}
\label{eq_20_equi}
\min_{\xvec,\svec}
\CSIM(\Hb\Db \svec,\yvec)+\alpha\norm \svec \norm_1+ \gamma \norm \Hb\Db \svec-\yvec \norm_2^2
\end{equation}
where $\alpha_1$ and $\gamma$ are chosen such that Karush-Kuhn-Tucker (KKT) conditions are satisfied \cite{LagBertsekas}. Note that since $\CSIM$ is convex and uni-modal, finding the local minima of \eqref{eq_20} is sufficient to attain its global optimum.
Now, introducing auxiliary variables $\xvec=\Db \svec$ and $\zvec=\Hb \xvec-\yvec$ the optimization problem \eqref{eq_20_equi} would change to:

\begin{equation}
\label{eq_21}
\min_{\xvec,\svec,\zvec}  \CSIM(\zvec) + \alpha \norm \svec \norm_1  + \gamma \norm \zvec \norm_2^2 \quad \text{s.t.}\quad \left\{\!\!\begin{array}{ll}
\xvec=\Db \svec \\
%\norm \zvec \norm <\epsilon_n \\
\zvec=\Hb \xvec-\yvec
\end{array}\right.
\end{equation}
The auxiliary variables are used to separate the main problem into simpler sub-problems.
Since the CSIM function is convex, it is guaranteed to use the Alternating Direction Method of Multipliers (ADMM) \cite{BoydADMM} to solve \eqref{eq_16}. Hence, the final cost function to be optimized is the augmented Lagrangian function:
\begin{align}
\label{eq_22}
\min_{\xvec,\svec,\zvec}  \, \lag (\xvec,\svec,\zvec)&=  \CSIM(\zvec) + \alpha \norm \svec \norm_1+ \gamma \norm \zvec \norm_2^2 \nonumber \\
& + \mub_1^T (\xvec-\Db \svec) + \frac{\sigma_1}{2} \norm\xvec-\Db \svec \norm_2^2  \nonumber \\
& + \mub_2^T(\zvec - \Hb \xvec+\yvec) + \frac{\sigma_2}{2} \norm\zvec - \Hb \xvec+\yvec \norm_2^2
\end{align}
The ADMM alternatively minimizes \eqref{eq_22} with respect to each variable while assuming the other variables  fixed. Hence, at each iteration of the ADMM, the problem \eqref{eq_22} is split into three sub-problems as follows:
\subsection{$\xvec$ sub-problem:}
\label{subsec:x}
The augmented Lagrangian cost function with respect to $\xvec$ is a quadratic function. Hence, the optimization sub-problem associating with $\xvec$ at $t$-th iteration of the ADMM is:
\begin{align}
\label{eq_23}
\xvec^{(t+1)}&=\mathop{\mathrm{argmin}}_{\xvec}
\lag(\xvec) =\lag (\xvec,\svec^{(t)},\zvec^{(t)}) =\frac{1}{2} \xvec^T \Qb\xvec- {\bvec ^{(t)}}^T \xvec
\end{align}
where $\Qb=\sigma_2\Hb^T \Hb +\sigma_1 \Ib$ and ${\bvec ^{(t)}}=\sigma_1 \Db \svec^{(t)} -\mub_1^{(t)}+\Hb^T (\sigma_2(\zvec^{(t)}+\yvec)+\mub_2^{(t)})$. The solution to this problem is simply obtained by $\xvec^{(t+1)} = \left(\sigma_1 \Ib+\sigma_2\Hb^T \Hb \right)^{-1} \bvec^{(t)}$.
Now, since $\Hb \Hb^T=\Ib$, using Sherman-Morrison-Woodbury lemma \cite{Woodbury}, the solution to \eqref{eq_23} is simplified to:
\begin{equation}
\label{eq_25}
\xvec^{(t+1)} =\frac{1}{\sigma_1} \left(\Ib-\frac{\sigma_2}{\sigma_2+\sigma_1} \Hb^T \Hb \right) \bvec^{(t)}
\end{equation}
\subsection{$\svec$ sub-problem:}
\label{subsec:s}
The optimization sub-problem associating with $\svec$ is:
\begin{align}
\label{eq_26}
\svec^{(t+1)}=\mathop{\mathrm{argmin}}_{\svec} \lag(\svec) &= \lag (\xvec^{(t+1)},\svec,\zvec^{(t)})
\end{align}
Using the Majorization Minimization (MM) technique \cite{LangeMM}, as proposed in \cite{Daub04}, this problem is shown \cite{SAMPTA_Jav} to be equivalent to minimization of a surrogate cost function denoted by $\lag^S(\svec,\svec_0)$:
%Assume $\norm \Db \norm_2^2 \leq \lambda$, using the Majorization Minimization (MM) technique \cite{LangeMM}, we define a surrogate function similar to what is proposed in \cite{Daub04}:
%\begin{align}
%\label{eq_27}
%\lag^S(\svec,\svec_0)= & \, \frac{1}{2}\norm\xvec^{(t+1)}-\Db \svec \norm_2^2 + \frac{1}{\sigma_1} {\mub_1^{(t)}}^T\! (\xvec^{(t+1)}-\Db \svec) \nonumber \\
%&+  \frac{\alpha_1}{\sigma_1} \norm \svec \norm_1 +  \frac{\lambda}{2} \norm \svec_0-\svec \norm_2^2 -\frac{1}{2} \norm\Db\svec_0-\Db \svec \norm_2^2
%\end{align}
%Since $\lag^S(\svec,\svec_0) \geq \lag(\svec), \, \forall \svec_0\neq \svec$ and $\lag^S(\svec,\svec) = \lag(\svec)$, optimizing \eqref{eq_27} with respect to $\svec$ will reduce the initial cost function $\lag(\svec)$. Hence by eliminating the unnecessary variables, the surrogate optimization problem is simplified to:
\begin{align}
\label{eq_28}
\min_{\svec}\lag^S(\svec,\svec_0) =&\min_{\svec} \,\frac{\lambda}{2} \norm \svec - \avec(\svec_0) \norm_2^2 + \frac{\alpha}{\sigma_1} \norm \svec \norm_1
\end{align}
where $\avec(\svec_0)=\svec_0+\frac{1}{\lambda } \Db^T \big(\xvec^{(t+1)}-\Db \svec_0+\frac{1}{\sigma_1}\mub_1^{(t)} \big)$ and $\lambda>\lambda_{\max}(\Db)$.
Let us set $\svec_0=\svec^{(t)}$, where $t$ denotes the iteration number. Now the solution to \eqref{eq_28} is obtained using the soft-thresholding operator and $\svec^{(t)}$ is updated according to:
\begin{align}
\label{eq_29}
\svec^{(t+1)} ={\cal S}_{\!{\frac{\alpha}{\lambda\sigma_1}} } \Big( \svec^{(t)}+\frac{1}{\lambda} \Db^T \big(\xvec^{(t+1)}-\Db \svec^{(t)}+\frac{1}{\sigma_1}\mub_1^{(t)}\big) \Big)
\end{align}
where ${\cal S}_{\tau}(\xvec)$ is a vector with component ${\cal S}_{\tau} (\xvec)_i$ obtained by:
\begin{equation}
\label{eq_13}
{\cal S}_{\tau} (\xvec)_i = {\cal S}_{\tau} (x_i)=\left\{\!\!\begin{array}{ll}
x_i,&  \vert x_i\vert >\tau \\
x_i - \tau,& \vert x_i \vert <\tau
\end{array}\right.
\end{equation}
\subsection{$\zvec$ sub-problem}
\label{subsec:z}
The sub-problem associating with $\zvec$ is as follows:
\begin{align}
\label{eq_30}
\zvec^{(t+1)}=\mathop{\mathrm{argmin}}_{\zvec} \lag(\zvec)&=\lag (\xvec^{(t+1)},\svec^{(t+1)},\zvec)\\
&=\CSIM(\zvec)+ {\mub_2^{(t)}}^T\!( \zvec - \Hb\xvec^{(t+1)}+\yvec ) \nonumber \\
 &\quad+\frac{\sigma_2}{2} \norm \zvec - \Hb\xvec^{(t+1)}+\yvec\norm_2^2 + \gamma \norm \zvec \norm_2^2\nonumber
\end{align}
Now substituting $\CSIM$ from \eqref{eq_1}, the resulting cost function will be in the quadratic form below:
\begin{align}
\label{eq_38}
\zvec^{(t+1)}&=\mathop{\mathrm{argmin}}_{\zvec} \frac{1}{2}\zvec^T\Kb \zvec- {\cvec^{(t)}}^T\!\!\zvec
\end{align}
where $\Kb= \sigma_2\Ib+ 2(\Wb+\gamma \Ib)=\zeta_1 \Ib_n + \zeta_2 \onevec_n \onevec_n^T$ with $\zeta_1 =\sigma_2+2 \frac{k_2}{n-1}+2\gamma$ and $\zeta_2=2\left(\frac{k_1}{n^2}-\frac{k_2}{n(n-1)}\right)$ and ${\cvec^{(t)}}=\sigma_2 (\Hb \xvec^{(t+1)}-\yvec)-\mub_2^{(t)}$. The solution to \eqref{eq_38} is obtained by:
\begin{equation}
\label{eq_z_update}
\zvec^{(t+1)}= \Kb^{-1}\cvec^{(t)}=\frac{1}{\zeta_1} \left(\Ib_n-\frac{\zeta_2}{\zeta_1+n \zeta_2} \onevec_n \onevec_n^T \right) \cvec^{(t)}
\end{equation}
where in the latter, we have applied the matrix inverse lemma.

%\begin{algorithm}[!b]
%
%\caption{CSIM-ALM algorithm with fixed $\alpha$ to address the optimization problem \eqref{eq_22}}
%\textbf{Set}   $\sigma_1,\, \sigma_2>0,\quad \alpha>0, \quad \gamma \geq 0, \quad \lambda>\lambda_{\max}(\Db)$,\newline $k_1, \,k_2>0$\newline
%\textbf{Initialize} $\mub_{1}^{(0)}=\mub_{2}^{(0)}=\zerovec$, $\zvec^{(0)}=\zerovec$, $\svec^{(0)}=\zerovec$, $t=0$.
%
%
%\begin{algorithmic}[1]
%\label{Algorithm_0}
%\REPEAT
%\STATE Update $\bvec^{(t)}$ and $\xvec^{(t+1)}$ using \eqref{eq_23} and \eqref{eq_25}
%\STATE Update $\svec^{(t+1)}$ using \eqref{eq_29}.
%\STATE Update $\zvec^{(t+1)}$  \eqref{eq_z_update}
%\STATE Update $\mub_{1}^{(t+1)}$ and $\mub_{2}^{(t+1)}$ according to \eqref{eq_40}
%\STATE $t \leftarrow t+1$
%\UNTIL {A stopping criterion is reached}
%\end{algorithmic}
%\end{algorithm}

\begin{algorithm}[!b]
\caption{CSIM-ALM algorithm with $\alpha$ continuation and backtracking.}
\textbf{Set}   $\sigma_1,\, \sigma_2>0,\quad \gamma\geq 0, \quad \beta>1,\quad \lambda>\lambda_{\max}(\Db),\quad \eta<1$, $\alpha_{\min}\ll1\quad k_1, \,k_2>0$, \newline
\textbf{Initialize} $\mub_{1}^{(0)}=\mub_{2}^{(0)}=\zerovec$, $\zvec^{(0)}=\zerovec$, $\svec^{(0)}={\svec^\ast}^{(-1)}=\zerovec$, $\alpha=\xi \norm\Db^T \yvec \norm_\infty$, $t=0$.

\begin{algorithmic}[1]
\label{Algorithm_1}
\REPEAT
\STATE Update $\bvec^{(t)}$ and $\xvec^{(t+1)}$ using \eqref{eq_23} and \eqref{eq_25}
\STATE Projection: $\xvec^{(t+1)}=(\Ib-\Hb)\xvec^{(t+1)}+\Hb\yvec$
\REPEAT
\STATE
Obtain ${\svec^\ast}^{(t)}$ by solving \eqref{eq_28} assuming $\svec_0=\svec^{(t)}$

\STATE
Set $\lambda = \lambda\times \beta$,
\UNTIL {$\lag({\svec^\ast}^{(t)}) \leq \lag^S({\svec^\ast}^{(t)}, \svec^{(t)})$}
\STATE Update $\svec^{(t+1)} = {\svec^\ast}^{(t)}$
\STATE Update $\zvec^{(t+1)}$  \eqref{eq_z_update}
\STATE Update $\mub_{1}^{(t+1)}$ and $\mub_{2}^{(t+1)}$ according to \eqref{eq_40}
\STATE Update $\alpha =\max\{\eta \alpha, \alpha_{\min}\} $
\STATE $t \leftarrow t+1$
\UNTIL {A stopping criterion is reached}
\end{algorithmic}
\end{algorithm}

\subsection{Multipliers update}
The final step of the ADMM is to update the Lagrangian multipliers associated with the equality constraints. Hence, we have:
\begin{align}
\label{eq_40}
\mub_1^{(t+1)}&=\mub_1^{(t)}+\sigma_1\big(\xvec^{(t+1)}-\Db \svec^{(t+1)}\big) \nonumber \\
\mub_2^{(t+1)}&=\mub_2^{(t)}+\sigma_2\big(\zvec^{(t+1)}-\Hb \xvec^{(t+1)}+\yvec \big)
\end{align}

%After all, the proposed iterative algorithm is given in Alg. \ref{Algorithm_0}.
\begin{theorem}
\label{theo:4}
If $\lambda>\lambda_{\max}(\Db)$ is satisfied and the parameters of CSIM, i.e., $k_1,k_2$ are positive, then, the proposed algorithm is guaranteed to converge to the optimal solution of problem \eqref{eq_20_equi} for fixed positive $\alpha$, $\sigma_1$, $\sigma_2$ and $\gamma$
\end{theorem}
\begin{IEEEproof}
The proof of convergence is given in Appendix \ref{section:App_Proof}.
\end{IEEEproof}

%\begin{figure}[!b]
%\centering
%\captionsetup[subfigure]{labelformat=empty}
%\centering
%\subfloat[]{\includegraphics*[width=0.8in]{Barbara256.png}}
%\subfloat[]{\includegraphics*[width=0.8in]{lena256.png}}
%\subfloat[]{\includegraphics*[width=0.8in]{camera256.png}}
%\subfloat[]{\includegraphics*[width=0.8in]{clown256.png}} \\[-5ex]
%\subfloat[]{\includegraphics*[width=0.8in]{Couple256.png}}
%\subfloat[]{\includegraphics*[width=0.8in]{house256.png}}
%\subfloat[]{\includegraphics*[width=0.8in]{orca256.png}}
%\subfloat[]{\includegraphics*[width=0.8in]{Peppers256.png}}
%\caption{The set of gray-scale images from which $8\times 8$ patches are extracted for simulations in this paper. From top to bottom and left to right: Barbara, Lena, Cameraman, Clown, Couple, House, Orca and Peppers}
%\label{fig_images}
%\end{figure}
\subsection{Remark}
\label{sec:remark}
We can enhance the performance of the proposed method using two techniques, namely backtracking and continuation. First of all, since $\svec^{(t+1)}$ is the minimizer of $\lag^S(\svec,\svec^{(t)})$, we have:
\begin{equation}
\label{eq_29a}
\lag(\svec^{(t+1)}) \leq \lag^S(\svec^{(t+1)},\svec^{(t)}) \leq \lag^S(\svec^{(t)},\svec^{(t)})=\lag(\svec^{(t)})
\end{equation}
where the first inequality implies from the fact that the surrogate function is the majorization of the original cost function. To ensure this condition is fulfilled, we may use a backtracking procedure to choose the appropriate value of $\lambda$. This method as proposed in \cite{Beck09}, solves the optimization problem \eqref{eq_28} and checks whether the solution $\svec^\ast$ satisfies $\lag(\svec^\ast) \leq \lag^S(\svec^\ast,\svec^{(t)})$. If true, the value of $\svec^{(t+1)}$ is set to $\svec^\ast$ and if not, it multiplies the value of $\lambda$ by a constant $\beta>1$. We may also improve the rate of convergence, using continuation on the regularizing parameter $\alpha$, i.e., we use a decreasing sequence for $\alpha^{(t)}$ instead of a fixed regularizing parameter. To this purpose, we choose exponentially decreasing $\alpha^{(t)}$ which alike \cite{Marv12}, results in an adaptive approach to specify the threshold $\frac{\alpha}{\lambda\sigma_1}$.
We also substitute corresponding samples of the reconstructed image $\xvec^{(t+1)}$ with known samples of the observed signal at each iteration.
There is of course a minimum threshold for $\alpha^{(t)}$ specified by the precision tolerance of the final solution. Thus, the enhanced version of the proposed method with $\alpha$ continuation is shown in Alg. \ref{Algorithm_1}.

\begin{table*}[!t]
\centering
\caption{Performance Comparison of the FIR denoising filters of order $m$ for simulations in part \ref{subsec:Exp1} (Input SNR=$1\mathrm{dB}$)}
\begin{tabular}{l|l||p{1.4cm}|p{1.4cm}|p{1.4cm}||p{1.4cm}|p{1.4cm}|p{1.4cm}}

%\multicolumn{1}{|c|}{\textbf{Animal}}
  \multicolumn{2}{c||}{}
  & \multicolumn{3}{c||}{$m=6$}
  & \multicolumn{3}{c}{$m=12$}\\[1	ex] \cline{3-8}
  \multicolumn{2}{c||}{}  & & & & & & \\[-0.5em]
  \multicolumn{2}{c||}{} & MSE Filter &CSIM Filter &SSIM Filter &MSE Filter &CSIM Filter &SSIM Filter \\ [-0.5em]
  \multicolumn{2}{c||}{}  & & & & & & \\\hline\hline

\multirow{4}{*}{Peppers} &PSNR (dB) & 21.64793 & \bf 21.7143 & 21.5236   & 21.49739 & \bf 21.5663 & 21.547 \\
   &SSIM &0.483457 & \bf 0.487075 & 0.48695  &0.541758 & \bf 0.547965 & 0.545625   \\
   &FSIM &0.737972 &\bf 0.743853 & 0.74152  &0.760147 & \bf 0.762156 & 0.759996   \\
   &Runtime (s)  & 8.84375 &\bf 8.671875 & 123.4844   & \bf 12.5 &12.90625  & 126.6406  \\ \hline

\multirow{4}{*}{Lena} &PSNR (dB) & 23.4301 & \bf 23.51699 & 23.5393   & 23.27777 & 23.29914 &\bf 23.31125  \\
   &SSIM &0.509773 &\bf 0.51945 & 0.516033   &0.554572 & \bf 0.559539 & 0.55748   \\
   &FSIM &0.76562 &\bf 0.770744 & 0.769893   &0.779354 &\bf 0.780314 & 0.780307  \\
   &Runtime (s) &\bf 8.546875 & 8.875  & 121.7813  & \bf 12.57813 &12.73438  & 125.0469  \\ \hline

\multirow{4}{*}{Barbara} &PSNR (dB) &22.07703 &\bf 22.09299 & 22.06136  &\bf 21.86944 & 21.86258 & 21.80966 \\
   &SSIM &0.471681 & \bf 0.474548 & 0.47164   &0.478401 & \bf 0.479339 & 0.473757 \\
   &FSIM &0.757897 &\bf  0.759037 & 0.755325   &0.760586 & \bf 0.763256 & 0.756748  \\
   &Runtime (s) &\bf  8.609375 &9.09375  & 122.1875   & \bf 12.51563 & 13.15625  & 125.1719  \\ \hline

\multirow{4}{*}{House} &PSNR (dB) & 23.43906 & \bf 23.5731 & 23.52321  & 24.12948 & \bf 24.16865 & 24.10161   \\
   &SSIM &0.386421 & \bf 0.394148 & 0.390097   &0.476552 & \bf 0.48155 & 0.479555    \\
   &FSIM &0.693577 & \bf 0.695039 & 0.694965   &0.731814 & \bf 0.733746 & 0.732695   \\
   &Runtime (s) & \bf 8.6875 & 9.09375  & 124.75   & \bf 12.64063 & 13.20313  & 129.3281   \\ \hline

\multirow{4}{*}{Cameraman} &PSNR (dB) &22.06733 & \bf 22.07957 & 22.05864   & 22.0714 & 22.06688 & \bf 22.10787  \\
   &SSIM &0.382708 & 0.396937 & \bf 0.39702    &0.453836 &\bf  0.457089 & 0.454604  \\
   &FSIM &0.674499 & \bf 0.683098 & 0.681842   &0.687971 & \bf 0.693188 & 0.689301  \\
   &Runtime (s)  & 8.96875 &\bf 8.921875 & 125.5   & \bf 12.75 &13.14063  & 123.0781  \\ \hline

\multirow{4}{*}{Couple} &PSNR (dB) & \bf 22.70483 & 22.67239 & 22.69083   &22.37438 & 22.37977 & \bf 22.38362  \\
   &SSIM &0.46712 & 0.467596 & \bf 0.470518  &\bf  0.458069 &0.457024 & 0.454858  \\
   &FSIM &0.758213 & 0.75606 & \bf 0.758453   &0.730827 & \bf 0.732256 & 0.728856   \\
   &Runtime (s) & \bf 8.625 & 9.25  & 124.0313   & \bf 12.29688 & 13.42188  & 125.4844   \\ \hline

\end{tabular}
\label{Table_2}
\end{table*}

\section{Simulation Results}
\label{sec:Simulations}
\subsection{Denoising}
\label{subsec:Exp1}
In this part, we conduct an experiment to show the performance of the proposed quality assessment criterion compared to some popular criteria, namely MSE, SSIM and FSIM \cite{FSIM11}. Consider $\xvec \in \Real ^N$ is the reference image signal and $\yvec \in \Real^N$ denotes the noisy observed image, i.e., $\yvec=\xvec+\nvec$ where $\nvec \in \Real^N$ is the noise signal with distribution $\nvec \sim \Gaus (0,\sigma_n^2)$. Suppose that the image is divided into small patches of size $\sqrt{n}\times\sqrt{n}$. The problem is to find a linear denoising filter $\hvec_j \in \Real^m (m\leq n)$ whose convolution with the $j$th patch of the image denoted by $\yvec_j \in \Real^n$ gives an estimate of the original signal denoted by $\hat{\xvec}_j$. i.e.
\begin{equation}
\label{eq_41}
\hat{x}_j[i] = \sum_{k=0}^{m-1} h_j[k] y_j[i-k], \quad i=0,1,\ldots, n-1
\end{equation}
After the small patches are denoised, the entire image is then reconstructed by superposition of the recovered patches.

In this scenario, the original image signal is unknown. In fact, $\xvec$ is the vector of spatial samples of a random process $x[i]$, which is assumed to be ergodic and WSS stationary. The clean image is then perturbed by a white Gaussian noise process $n[i]$ with variance $\sigma_n^2$, whose samples are denoted by $\nvec$. The observed noisy signal $\yvec$ is thus, modelled by the sum of theses two processes. Hence, in this case, the problem of finding the equalizer filter, is indeed a linear estimation problem as discussed in \cite{Chann08}. But, here we assume stationariness within the spatial domain of each patch. Consequently, we confine ourselves to local patch denoising for estimation of $\hvec_j$. The usual fidelity criterion for estimation is MSE, which leads to the so-called Wiener-Hopf \cite{Wiener2000} equations:
\begin{align}
\label{eq_45_1}
\hvec_j^{\mathrm{MSE}}\!\!= \!\mathop{\mathrm{argmin}}_{\hvec_j}\Expv \Big[ \Big(x_j[i]-\!\!\sum_{k=0}^{m-1} h_j[k] y_j[i-k] \Big)\!^2\Big]\!=\!\Rb_{y_j,y_j}^{-1}  \rvec_{x_j,y_j}
\end{align}
where $\Rb_{y_j,y_j}$ and $\rvec_{x_j,y_j}$ denote the auto-correlation matrix and the the vector of cross-correlation components respectively, i.e., $\Rb_{y_j,y_j}[k,l]=r_{y_j,y_j}[k-l], \quad \rvec_{x_j,y_j}[k]= r_{x_j,y_j}[k]$.
%\begin{align} \nonumber \\
%k,l=0,\ldots, m-1
%, \quad k=0, \ldots, m-1
%\end{align}
%\begin{align}
%\rvec_{x,y}&=\Expv [ x[0]\yvec] \\
%&= \cvec_{x,y}+\mu_y^2 \onevec \nonumber
%\end{align}
Now, assume the noise process is independent from $\xvec$ and is distributed homogeneously across the whole image. Using $\yvec_j=\xvec_j+\nvec_j, \, \nvec_j \bot \xvec_j$ with $\mu_{n_j}=0$ and $c_{n_j,n_j}[k]=\sigma_n^2 \delta[k] $, it can be shown that \cite{Chann08}
\begin{align}
\label{eq_45_cmu}
c_{x_j,y_j}[k]&= c_{y_j,y_j}[k] -\sigma_n^2 \delta[k], \quad \mu_{x_j} = \mu_{y_j}\\
r_{x_j,y_j}[k]&=c_{x_j,y_j}[k]+\mu_{y_j}^2, \quad r_{y_j,y_j}[k]=c_{y_j,y_j}[k]+\mu_{y_j}^2\nonumber 
\end{align}
Instead of the MSE criterion, we may use CSIM or SSIM in our estimation problem. If we use the statistical definition of CSIM, the optimization problem for finding the denoising filter $\hvec_j$ would be:
 \begin{align}
\label{eq_45_2}
\hvec_j^{\mathrm{CSIM}}= \mathop{\mathrm{argmin}}_{\hvec_j} k_1(\mu_{x_j}-\mu_{\hat{x}_j})^2+k_2(\sigma_{x_j}^2+\sigma_{\hat{x}_j}^2-2\sigma_{x_j,{\hat{x}_j}})
\end{align}
Substituting ${\hat{x}_j}$ from \eqref{eq_41} we have the following relations:
\begin{align}
\label{eq_45_3}
\mu_{\hat{x}_j} &= \Expv\Big[\sum_{k=0}^{m-1} h_j[k] y_j[i-k] \Big] =\sum_{k=0}^{m-1} h_j[k] \mu_{y_j} \nonumber\\
\sigma_{x_j,\hat{x}_j}&=\sum_{k=0}^{m-1}h_j[k] c_{x_j,y_j}[k]=\hvec_j^T \cvec_{x_j,y_j} 
\end{align}
If we denote the covariance matrix of $\yvec_j$ by $\Cb_{y_j,y_j}$, the optimization problem \eqref{eq_45_2} is simplified to:

\begin{align}
\label{eq_45_5}
 \hvec_j^{\mathrm{CSIM}} =\mathop{\mathrm{argmin}}_{\hvec_j }\, & k_1(\mu_{x_j}-\onevec_m^T \hvec_j \mu_{y_j})^2 + \\
 & k_2  \Big(\sigma_{x_j}^2 +\hvec_j^T \Cb_{y_j,y_j} \hvec_j - 2\hvec_j^T \cvec_{x_j,y_j} \Big) \nonumber
\end{align}
This problem is quadratic in terms of $\hvec_j$. To solve this, we use \eqref{eq_45_cmu} and we differentiate with respect to $\hvec_j$ which gives:
%\begin{align}
%\label{eq_45_6}
%-2 k_1 \mu_{y_j}^2(1-\onevec_m^T \hvec_j)  \onevec_m & + \\
% & 2k_2  \Big( \Cb_{y_j,y_j} \hvec_j -  \cvec_{x_j,y_j} \Big) = \zerovec\nonumber
%\end{align}
%Hence,
\begin{align}
\label{eq_45_7}
\hvec_j^{\mathrm{CSIM}} =  \Big(\Cb_{y_j,y_j}+\frac{k_1}{k_2} \mu_{y_j}^2 \onevec_m \onevec_m^T\Big) ^{-1} \Big(  \cvec_{x_j,y_j}+\frac{k_1}{k_2}\mu_{y_j}^2\onevec_m \Big) \nonumber
\end{align}
To reduce the complexity of calculating the inverse above, having $\Cb_{y_j,y_j}^{-1}$, we can use matrix inverse lemma.

Now, let us turn to SSIM fidelity criterion. The SSIM optimization problem for estimating the optimal equalization filter, tries to maximize the cost function below with respect to $\hvec_j$:
%\begin{align}
%\label{eq_46}
%\hvec_j^{\mathrm{SSIM}} &= \mathop{\mathrm{argmax}}_{\hvec_j } \\
\begin{equation}
\Big( \dfrac{2\mu_{y_j}^2 \hvec_j^T \onevec_m +C_1}{\mu_{y_j}^2(1+\hvec_j^T \onevec_m \onevec_m^T \hvec_j)+C_1} \Big) \Big( \dfrac{2\hvec_j^T \cvec_{x_j,y_j}+C_2}{\sigma_{x_j}^2+\hvec_j^T \Cb_{y_j,y_j}\hvec_j  +C_2} \Big)
\end{equation}
This optimization problem is solved via conversion to a quasi-convex cost, using bi-sectional search method \cite{Chann08}.

In practical simulations for denoising image patches, the covariance matrix $\Cb_{y_j,y_j}$ is empirically estimated using unbiased estimation and we use equation \eqref{eq_45_cmu} for the values of $\sigma_{x_j}$ and $\mu_{x_j}$. The simulation results for denoising with FIR filters are given in table \ref{Table_2}. It is clear that the proposed CSIM index is outperforming other criteria in terms of estimation quality at low SNR. It is even faster in several cases and generally more efficient than MSE and SSIM in the sense of FSIM performance as a tertiary index.

\subsection{Sparse recovery}
\label{subsec:Exp2}
In this experiment, we compare the quality performance of the proposed CSIM-ALM method for recovery of missing samples of image patches with some popular sparse recovery algorithms. We use IMATCS\footnote{{http://ee.sharif.edu/$\sim$imat/}}, L1-LS, DALM\footnote{{{https://people.eecs.berkeley.edu/$\sim$yang/software/l1benchmark/}}}, TV\footnote{{http://www.caam.rice.edu/$\sim$optimization/L1/TVAL3/}} \cite{TV013}, FISTA, SL0\footnote{{http://ee.sharif.edu/$\sim$SLzero/}}, GOMP\footnote{{http://islab.snu.ac.kr/paper/gOMP.zip}}, and the method in \cite{Oga13}, which we call SSIM-based Matching Pursuit (SSIM-MP).

For simulations of this part, we use $8\times8$ image patches vectorized using raster scanning, and we select 100 patch vectors of size $n=64$ denoted by $\xvec_i$ at random. For each patch, a binary random sampling mask $\Hb_j$ with $m_j$ ones, is generated and the observed image signal for each experiment is acquired by $\yvec_{i,j}=\Hb_j \odot \xvec_i$ where $\odot$ denotes pointwise (Hadamard) product. The locations of $1$s are chosen uniformly at random and the sampling ratio of the signal defined as $SR_j=\frac{m_j}{n}$ varies between $(0,1)$. We use complete ($64\times 64$) and over-complete ($64\times 128$) DCT and Haar Wavelet Packet (WP)\footnote{MATLAB commands \texttt{wmpdictionary} and \texttt{wphaar}} dictionaries for sparse representation.

\begin{figure*}[!t]
%\vspace{-1cm}
\centering
\hspace{-1cm}
\subfloat[PSNR]{\includegraphics[width=2in]{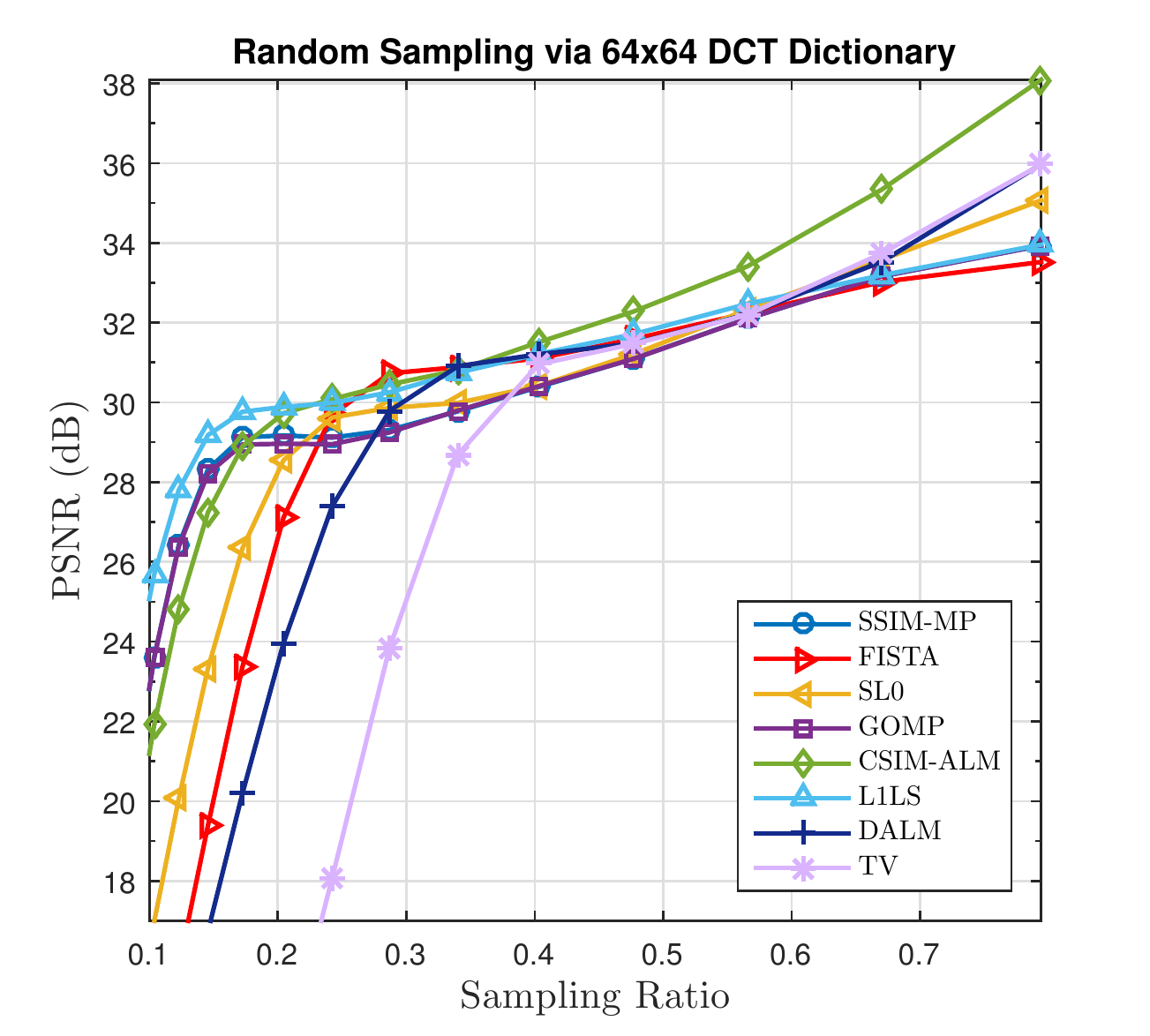}%
\label{fig2:a}}  \hspace{-0.55cm}
\subfloat[SSIM]{\includegraphics[width=2in]{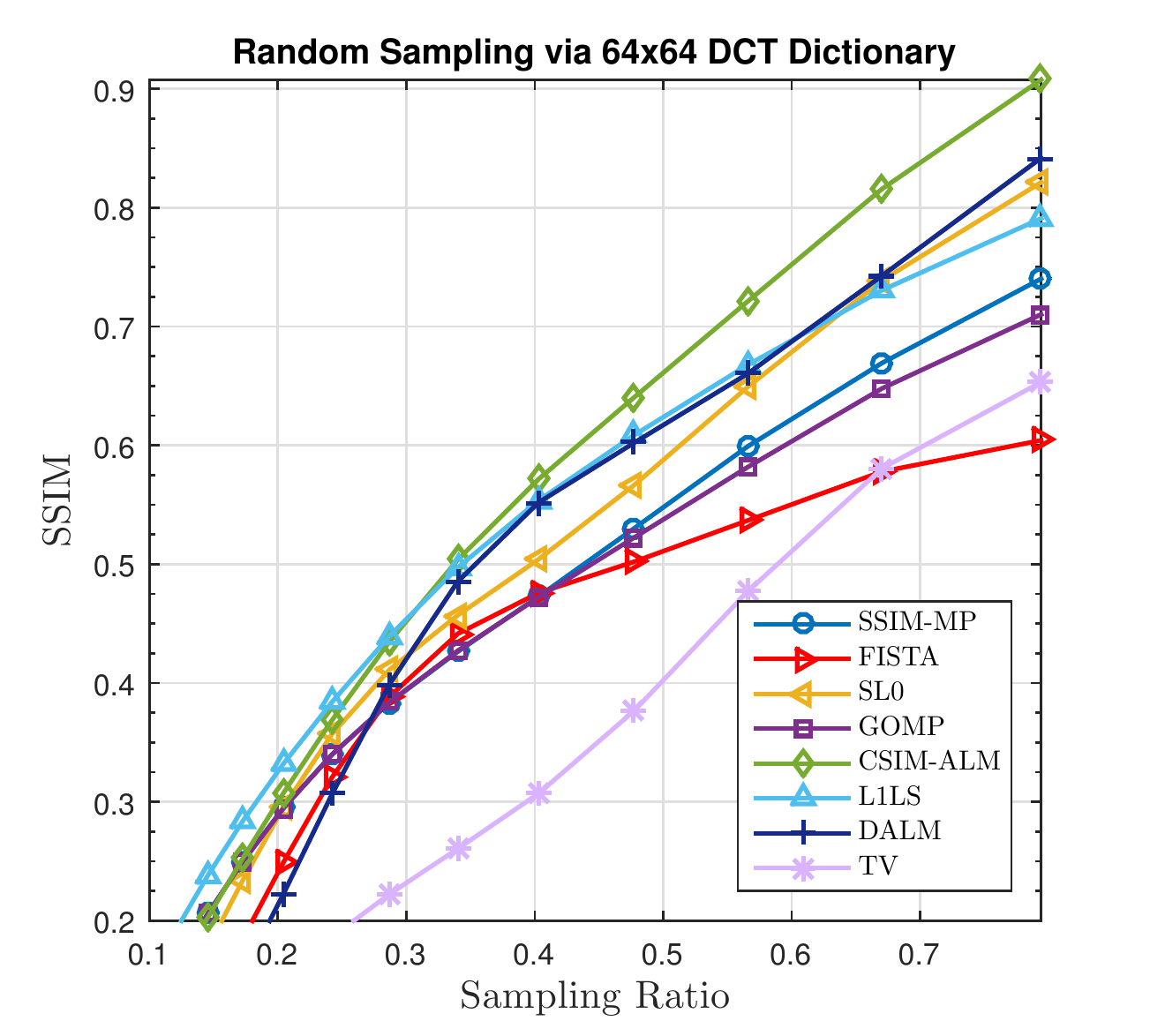}%
\label{fig2:b}} \hspace{-0.55cm}
\subfloat[PSNR]{\includegraphics[width=2in]{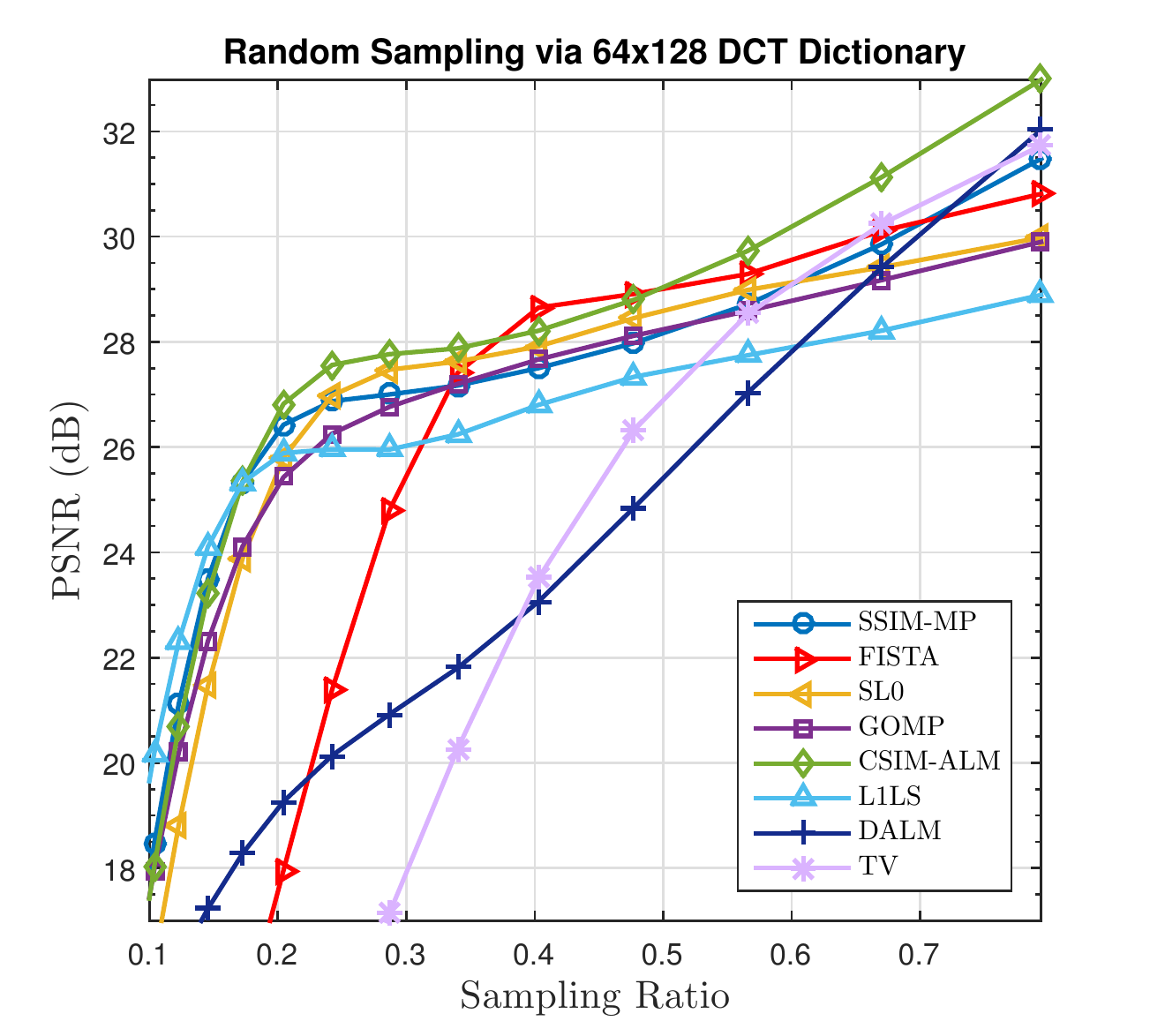}%
\label{fig2:c}}  \hspace{-0.55cm}
\subfloat[SSIM]{\includegraphics[width=2in]{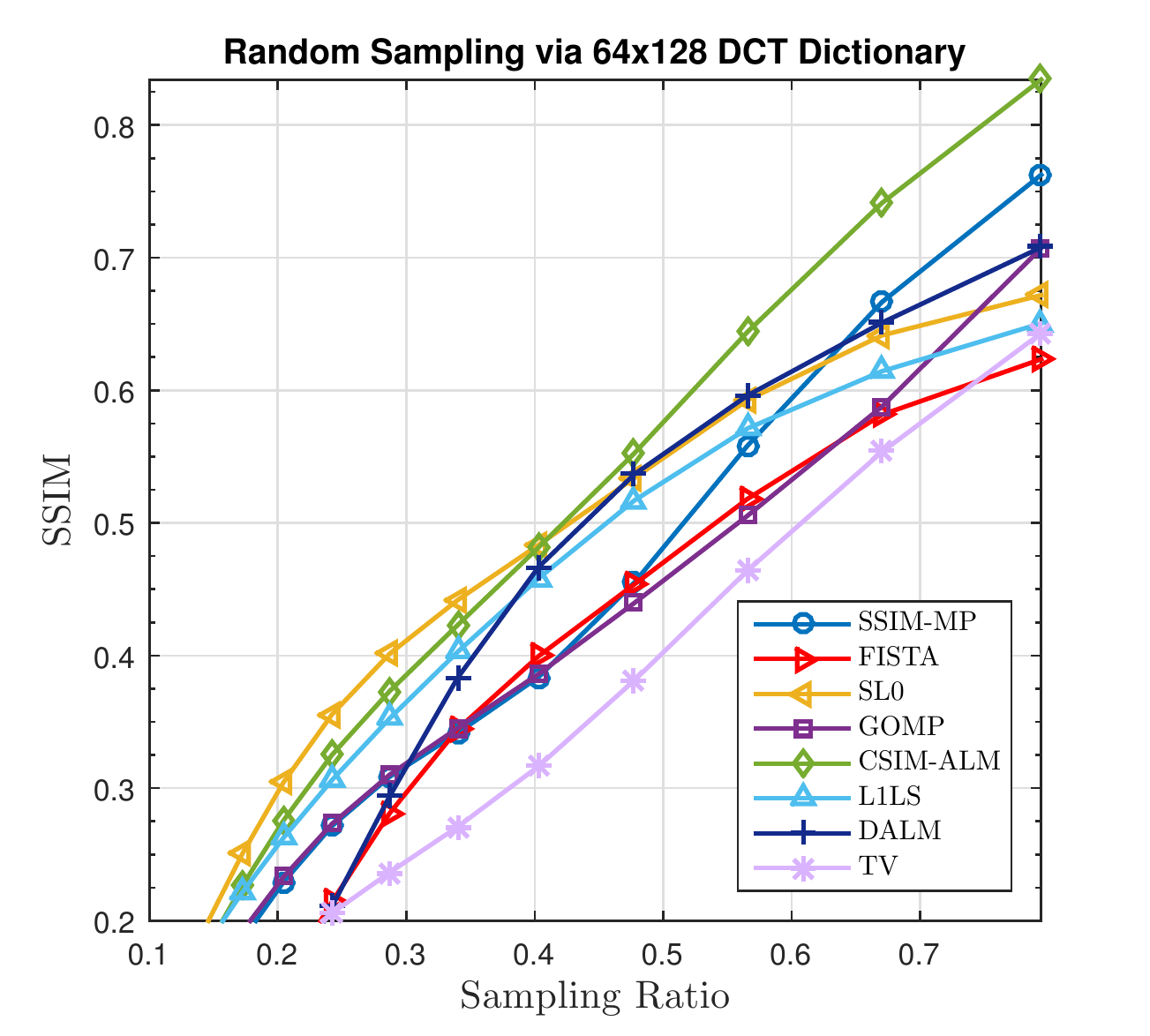}%
\label{fig2:d}}  \\[-2.5ex]
\hspace{-1cm}
\subfloat[PSNR]{\includegraphics[width=2in]{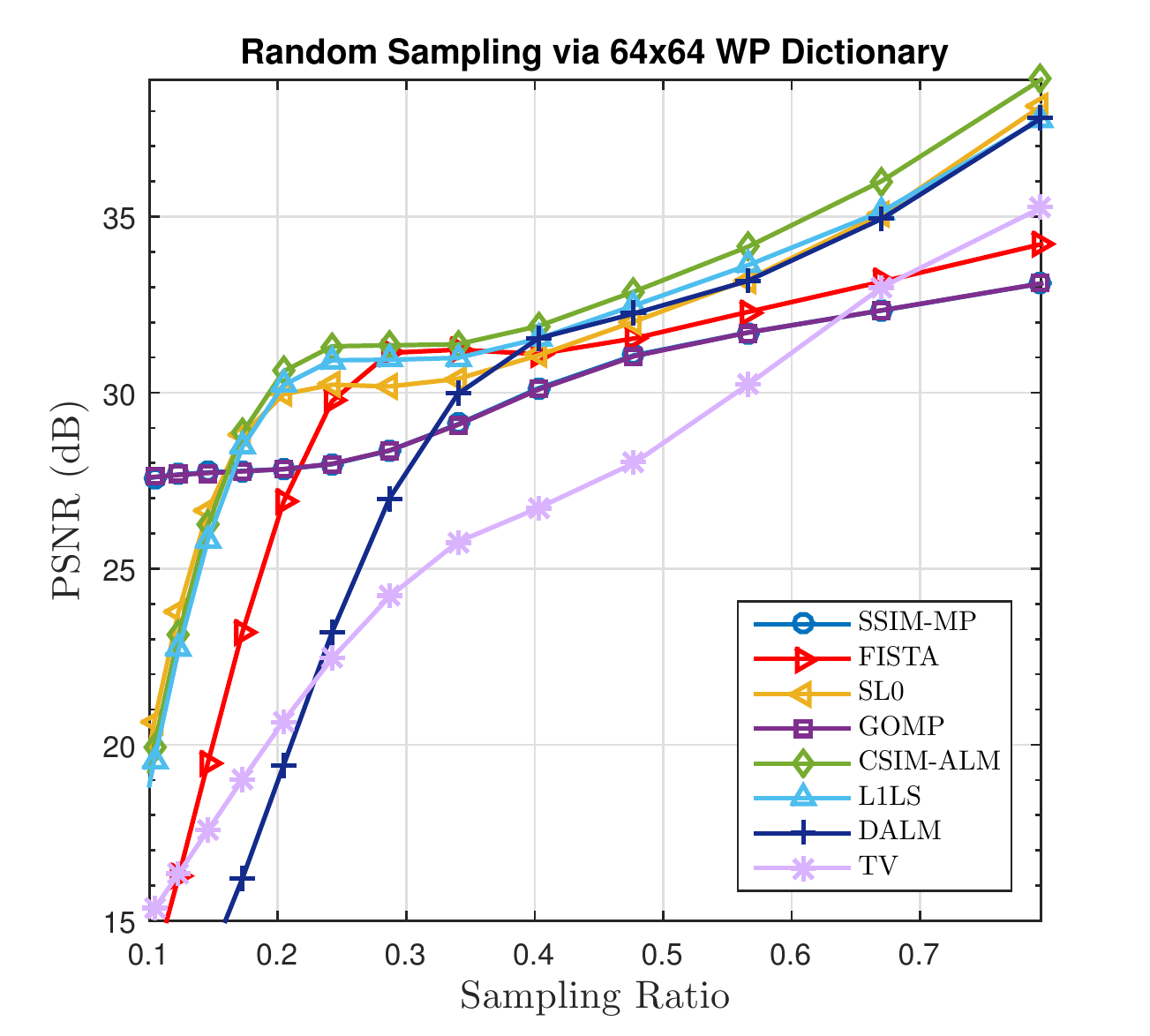}%
\label{fig2:e}}\hspace{-0.55cm}
\subfloat[SSIM]{\includegraphics[width=2in]{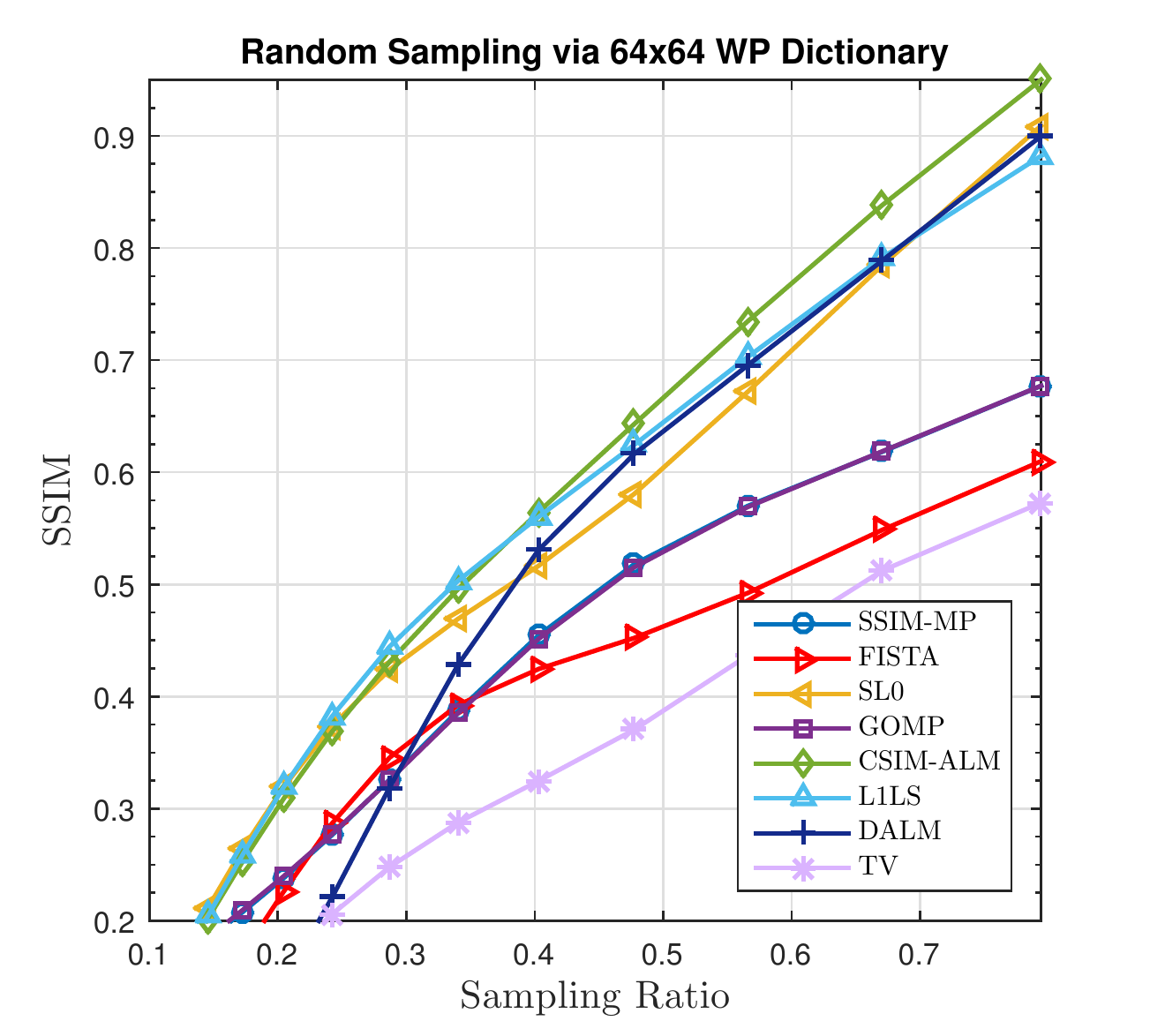}%
\label{fig2:f}}\hspace{-0.55cm}
\subfloat[PSNR]{\includegraphics[width=2in]{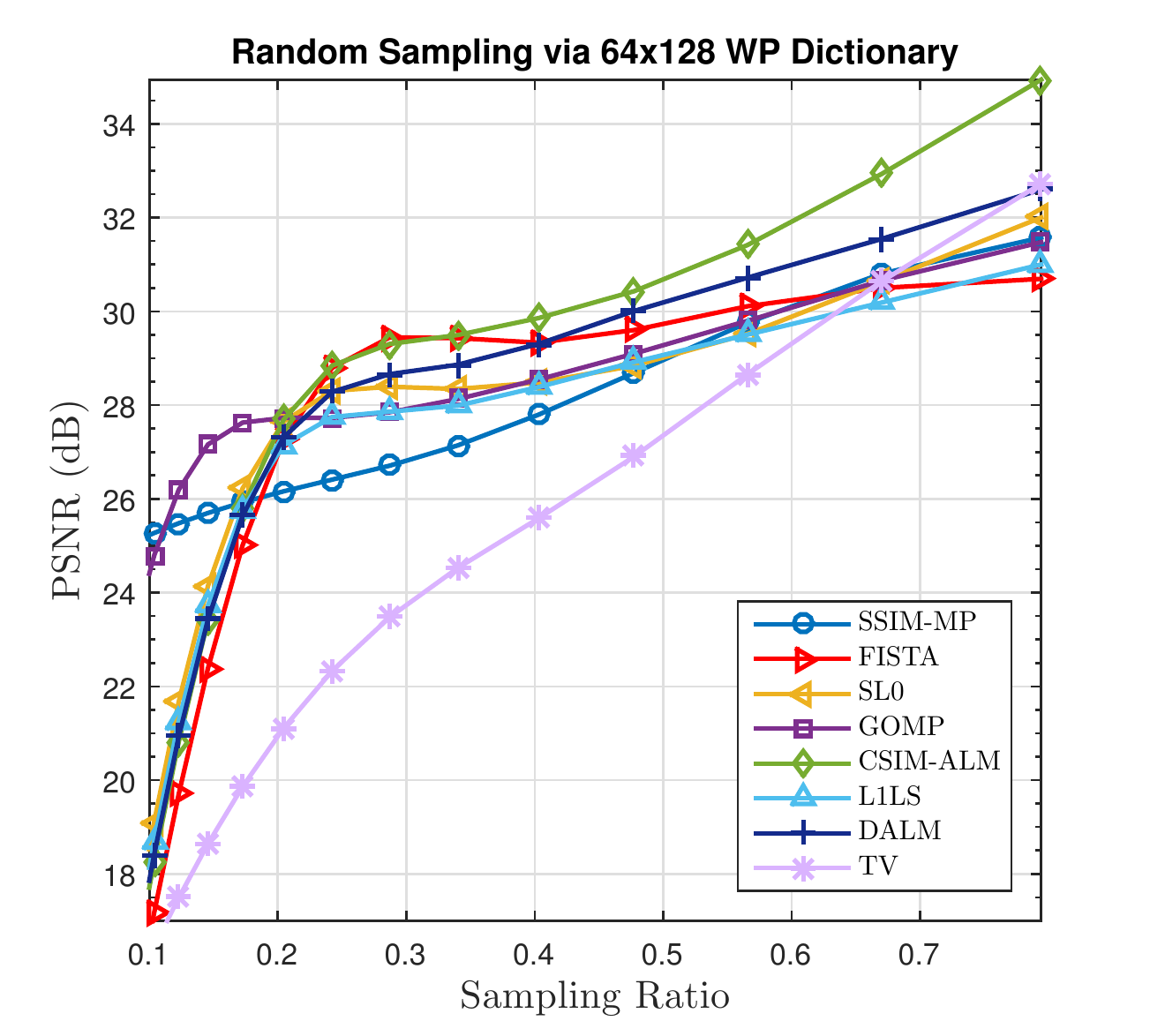}%
\label{fig2:g}}\hspace{-0.55cm}
\subfloat[SSIM]{\includegraphics[width=2in]{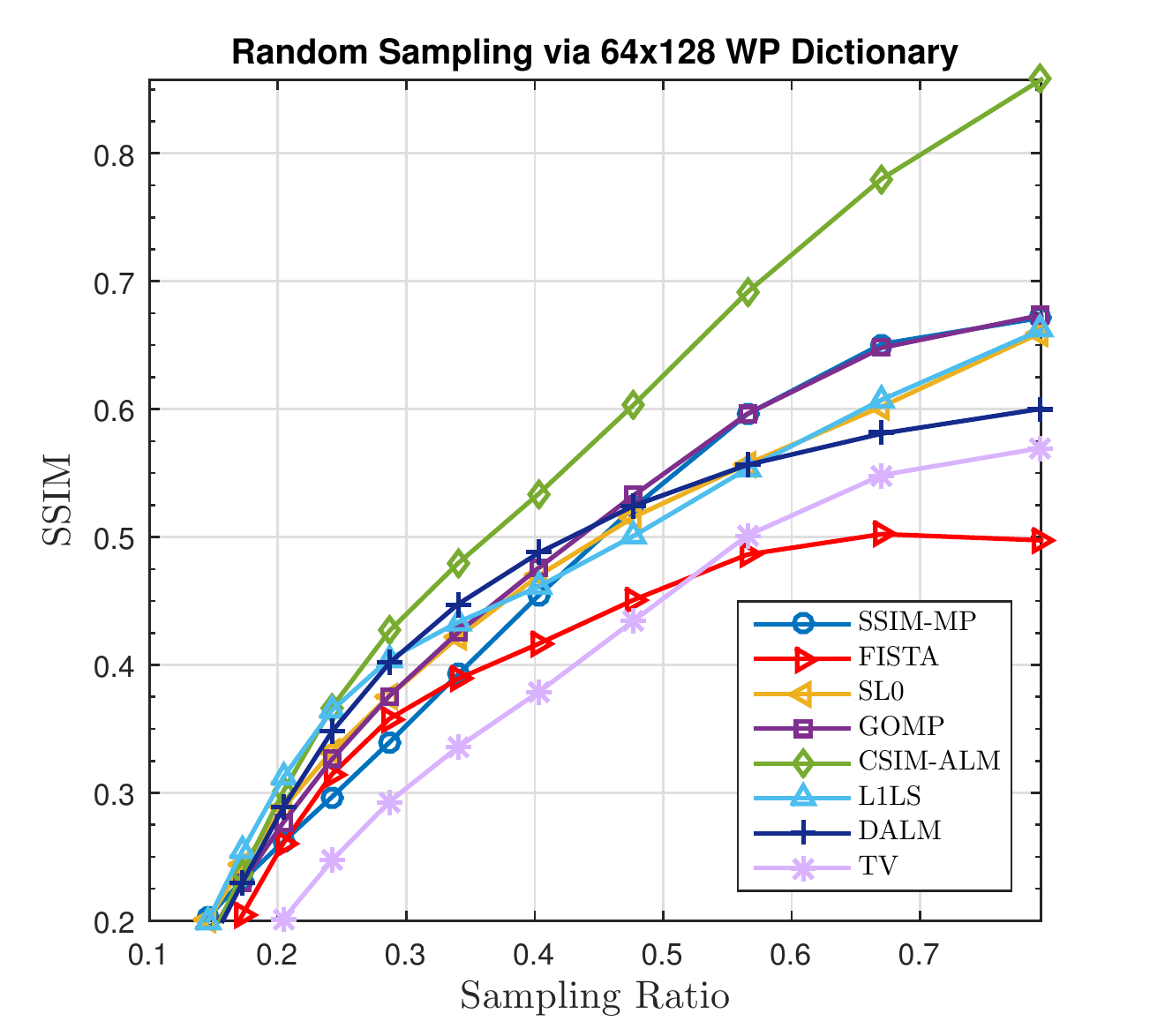}%
\label{fig2:h}}
\caption{Quality performance of sparse recovery methods versus the rate of random sampling of $64\times 1$ image vectors. We have assumed sparse approximation via: \ref{fig2:a} and \ref{fig2:b}) $64\times 64$ DCT dictionary,  \ref{fig2:c} and \ref{fig2:d}) $64\times 128$ DCT dictionary, \ref{fig2:e} and \ref{fig2:f}) $64\times 64$ Haar WP dictionary, \ref{fig2:g} and \ref{fig2:h}) $64\times 128$ Haar WP dictionary.}
\label{fig_2}
\end{figure*}

\begin{figure*}[!t]
\vspace{-0.4cm}
\centering
\subfloat[$sr=0.8$]{\includegraphics*[width=2.3in]{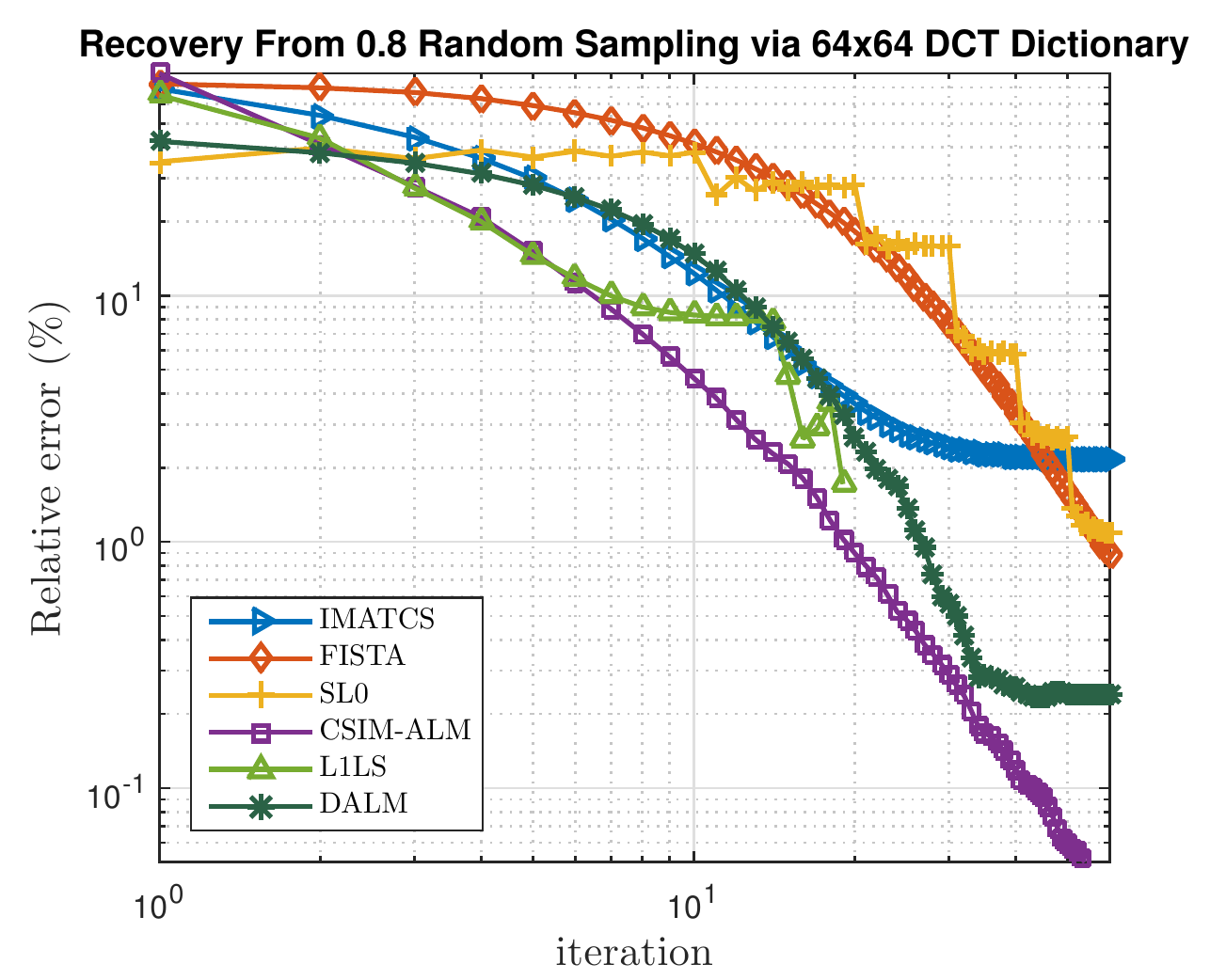}\label{fig_err:1}\hspace{-0.3cm}}
\subfloat[$sr=0.6$]{\includegraphics*[width=2.3in]{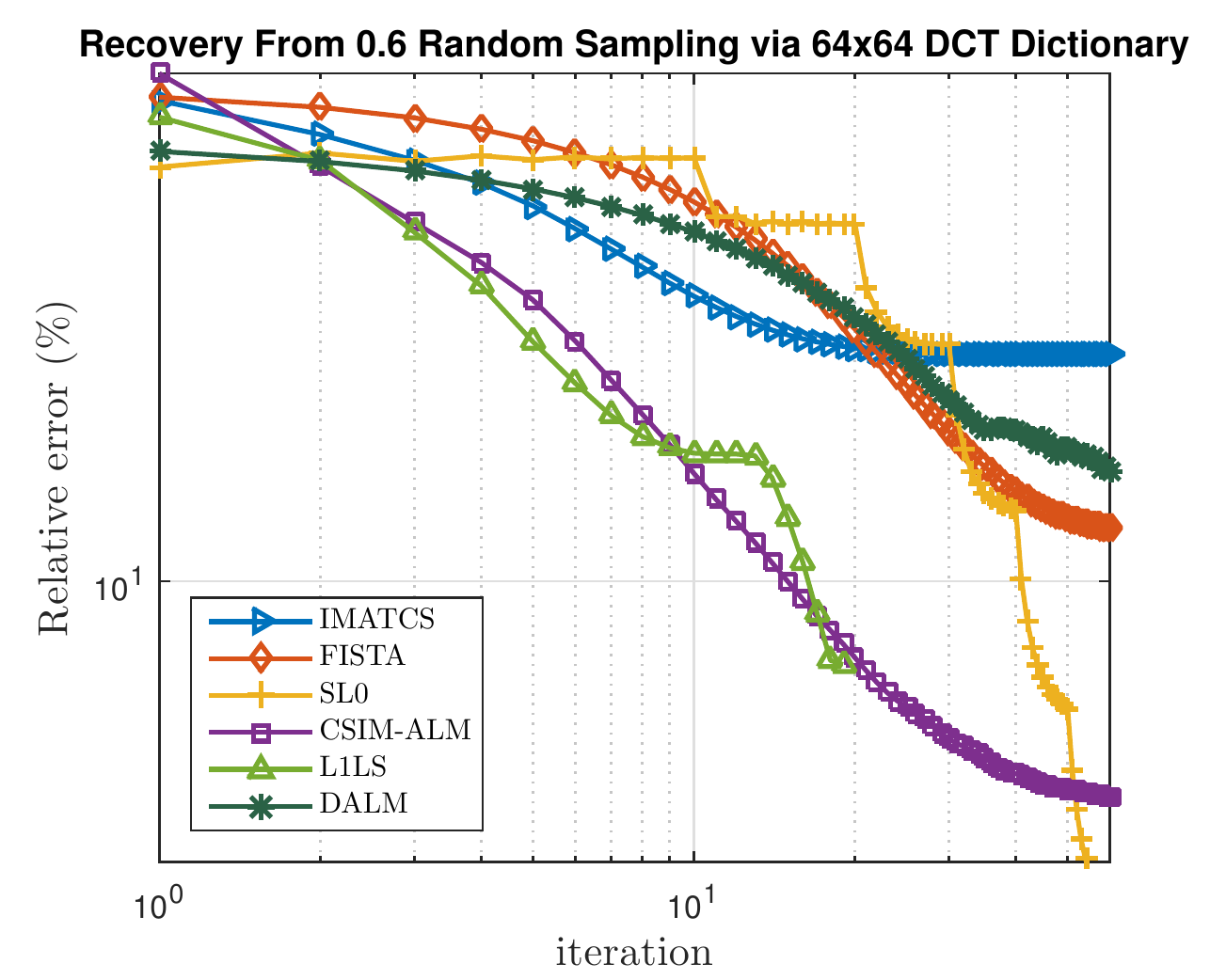}\label{fig_err:2}\hspace{-0.3cm}}
\subfloat[$sr=0.4$]{\includegraphics*[width=2.3in]{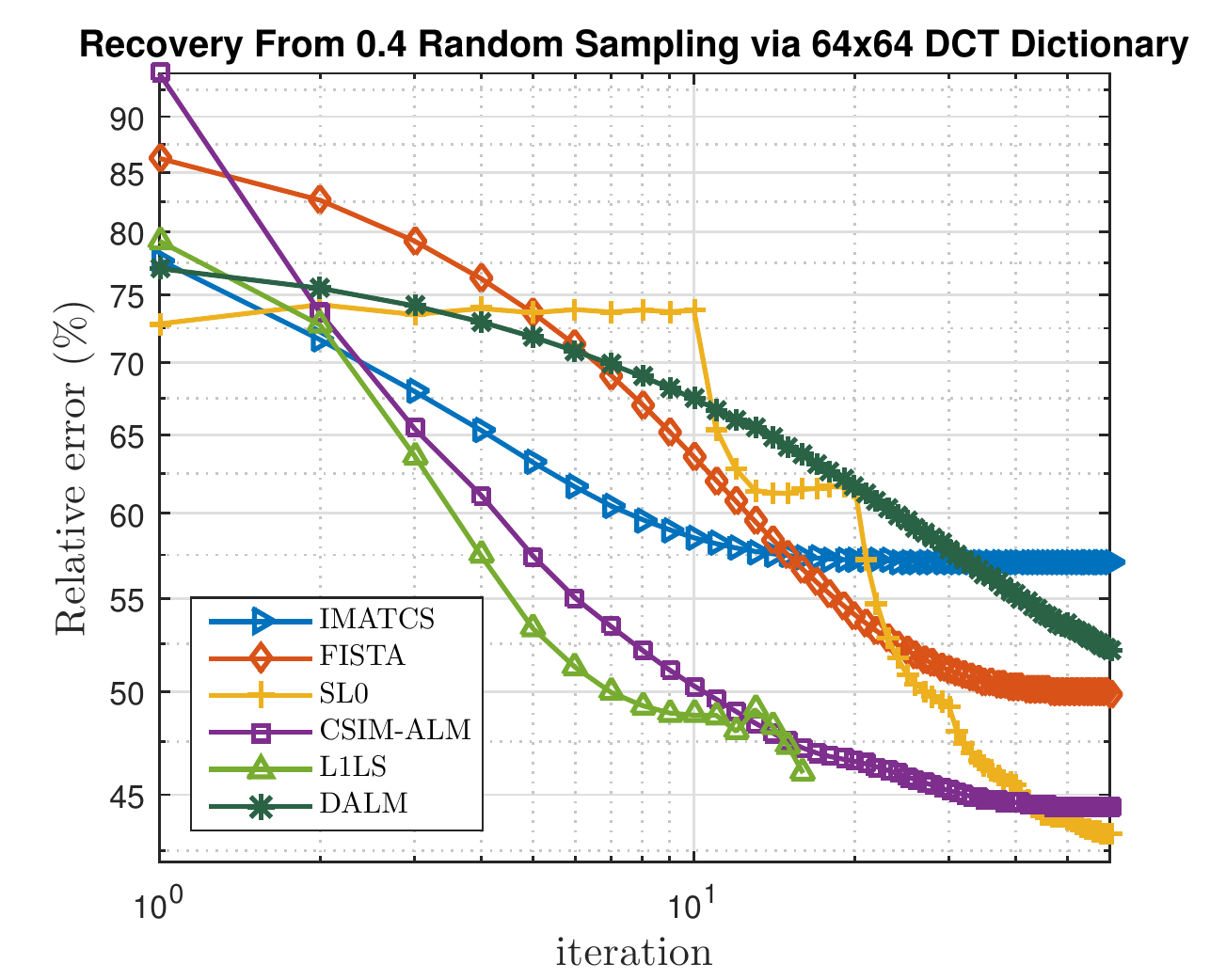}\label{fig_err:3}}  \\[-1.8 ex]
\subfloat[$sr=0.8$]{\includegraphics*[width=2.3in]{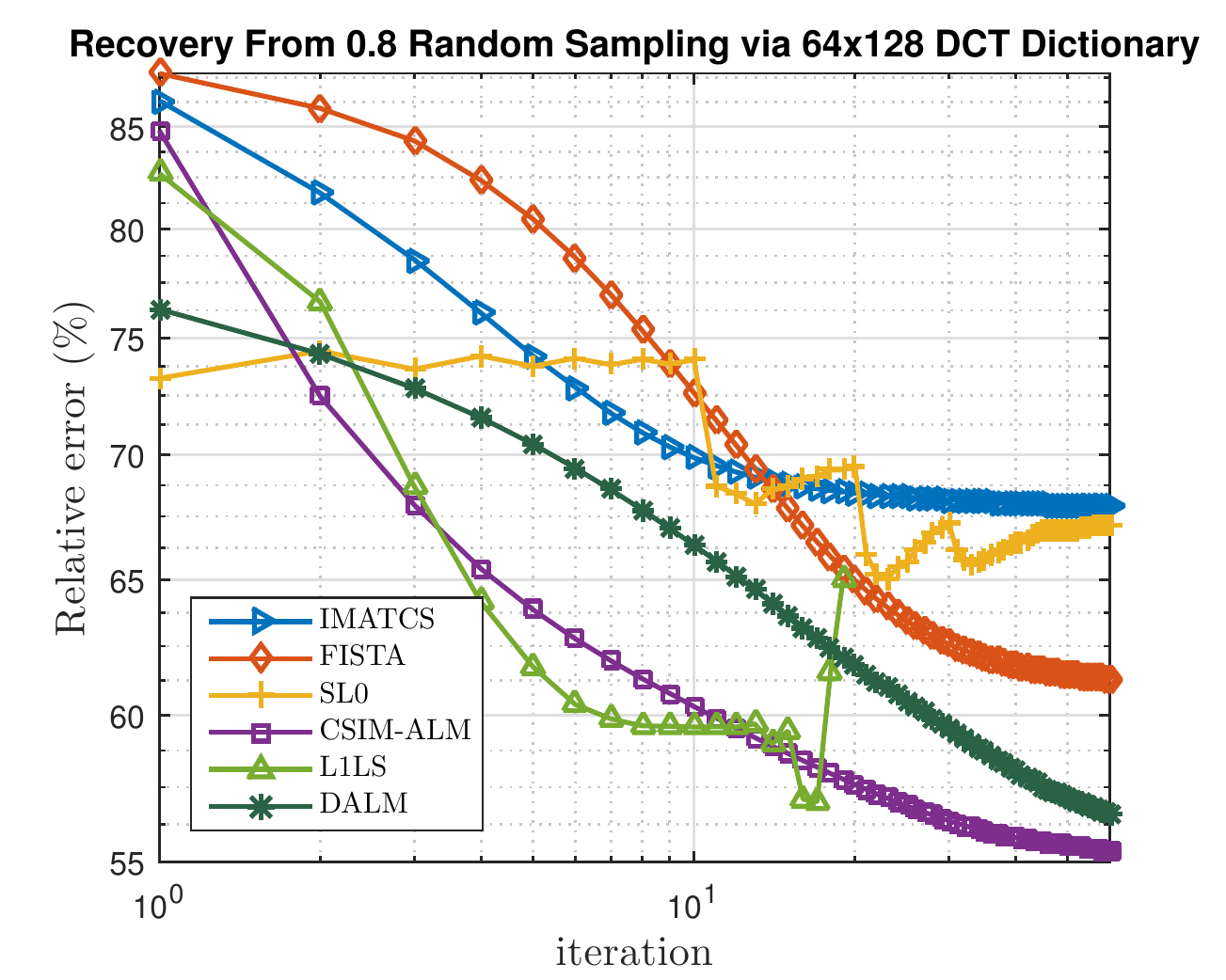}\label{fig_err:4}\hspace{-0.3cm}}
\subfloat[$sr=0.6$]{\includegraphics*[width=2.3in]{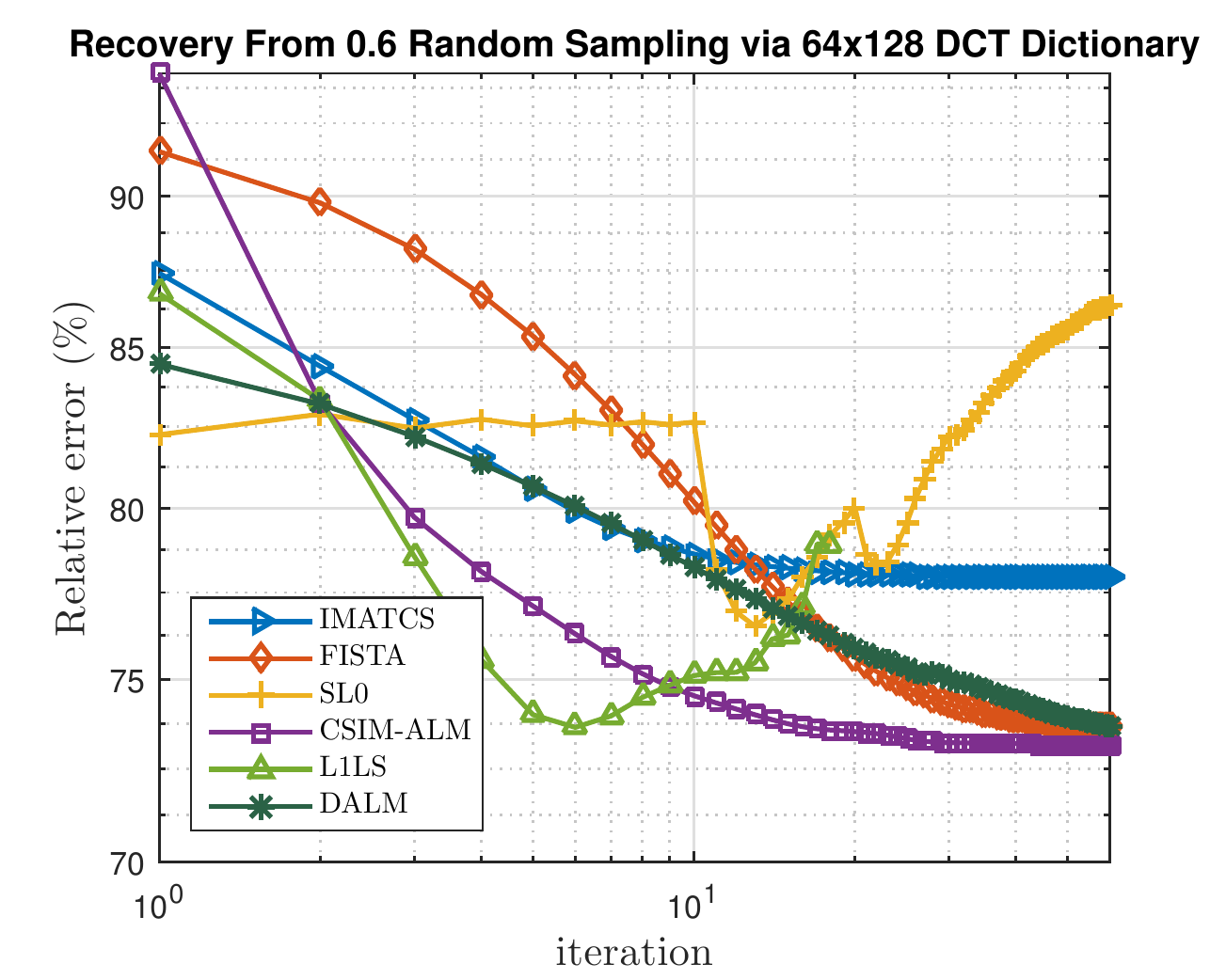}\label{fig_err:5}\hspace{-0.3cm}}
\subfloat[$sr=0.4$]{\includegraphics*[width=2.3in]{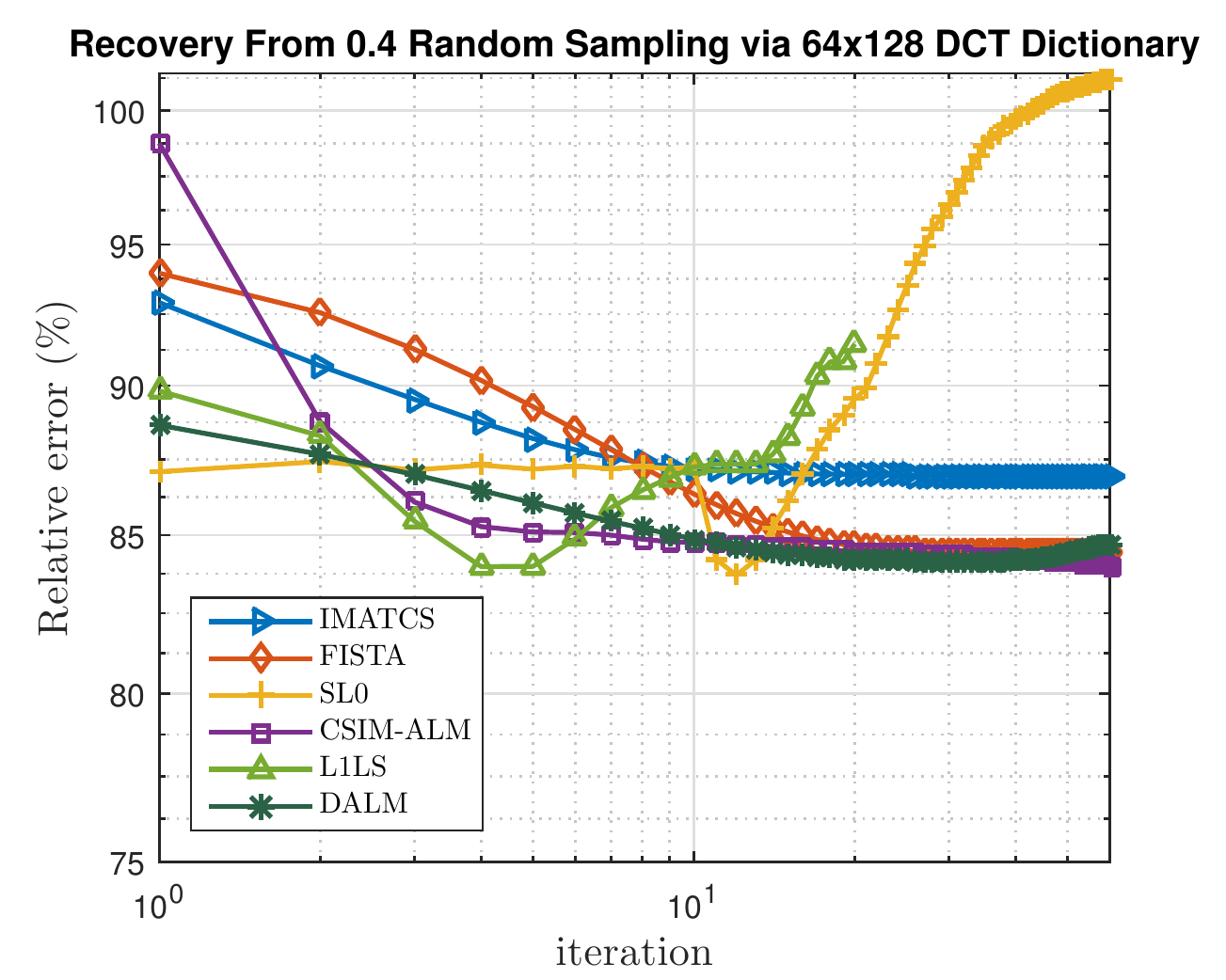}\label{fig_err:6}}
 \caption{Relative reconstruction error versus the number of iterations for sparse approximation via $64\times 64$ (\ref{fig_err:1} to \ref{fig_err:3}) and $64\times 100$ (\ref{fig_err:4} to \ref{fig_err:6}) DCT atoms.}
 \label{fig_err}
\end{figure*}

\subsubsection*{Natural image signals}
In this scenario, we extract $8\times8$ patches from natural grayscale images. Thus, the exact sparsity is unknown. Hence, to use matching pursuit algorithms, we consider $10\%$ sparsity. After recovery of missed samples, we then average over random experiments (random $\xvec_i$s and random masks with same $m_j$) and plot the PSNR and SSIM, versus the sampling rate. The parameters for SL0 and TV are set to their defaults. The values of the parameters for IMATCS are set to $\alpha=0.2$, $\beta = 0.5 \norm \Db^T \Hb^T\yvec \norm_\infty$, $\lambda_{\min}=1e-3$. The stopping criterion for DALM and L1-LS are set to their defaults, meaning that the algorithms stop when the duality gap falls below a certain tolerance. The stopping criterion for the remaining algorithms including IMATCS, FISTA, SL0 and CSIM-ALM is set to the maximum iteration count which is 50. The parameters for CSIM-ALM are chosen as $\sigma_1 = 0.4\frac{m}{n}$, $\sigma_2=2\frac{m}{n}$, $\gamma=1$, $\xi=0.1$, $\eta=0.95$, $\beta=1.1$, $\alpha_{\min}1e-4$, $k_2=n-1=63$ and $k_1 = 0.25k_2$. The choices for $k_1$ and $k_2$ are empirically obtained considering constraints in \ref{Corr1}. Fig.~\ref{fig_2} shows the results of the sparse recovery from random samples. The top row figures depict reconstruction via DCT atoms and the bottom row show the results via Haar Wavelet Packet dictionary. As shown in these figures, the proposed CSIM-ALM algorithm, mostly outperforms the state of the art methods for sparse recovery via \lone-norm minimization specifically at $SR>0.3$. It mainly provides a better reconstruction quality compared to DALM which commonly uses the ADMM technique to solve the \lone~optimization problem. This superiority is particularly more apparent in terms of SSIM performance. In fact, the proposed algorithm also outperforms SSIM-MP which is based on non-convex SSIM maximization.
\subsubsection*{Artificially generated signals}
\begin{figure}[!t]
%\vspace{-1cm}
\centering
%\subfloat[$sr=0.8$]{\includegraphics*[width=2.2in]{Time_64_sr_0_8-eps-converted-to.pdf}\label{fig_time:1}\hspace{-0.3cm}}
%\subfloat[$sr=0.6$]{}
\includegraphics[width=2.5in]{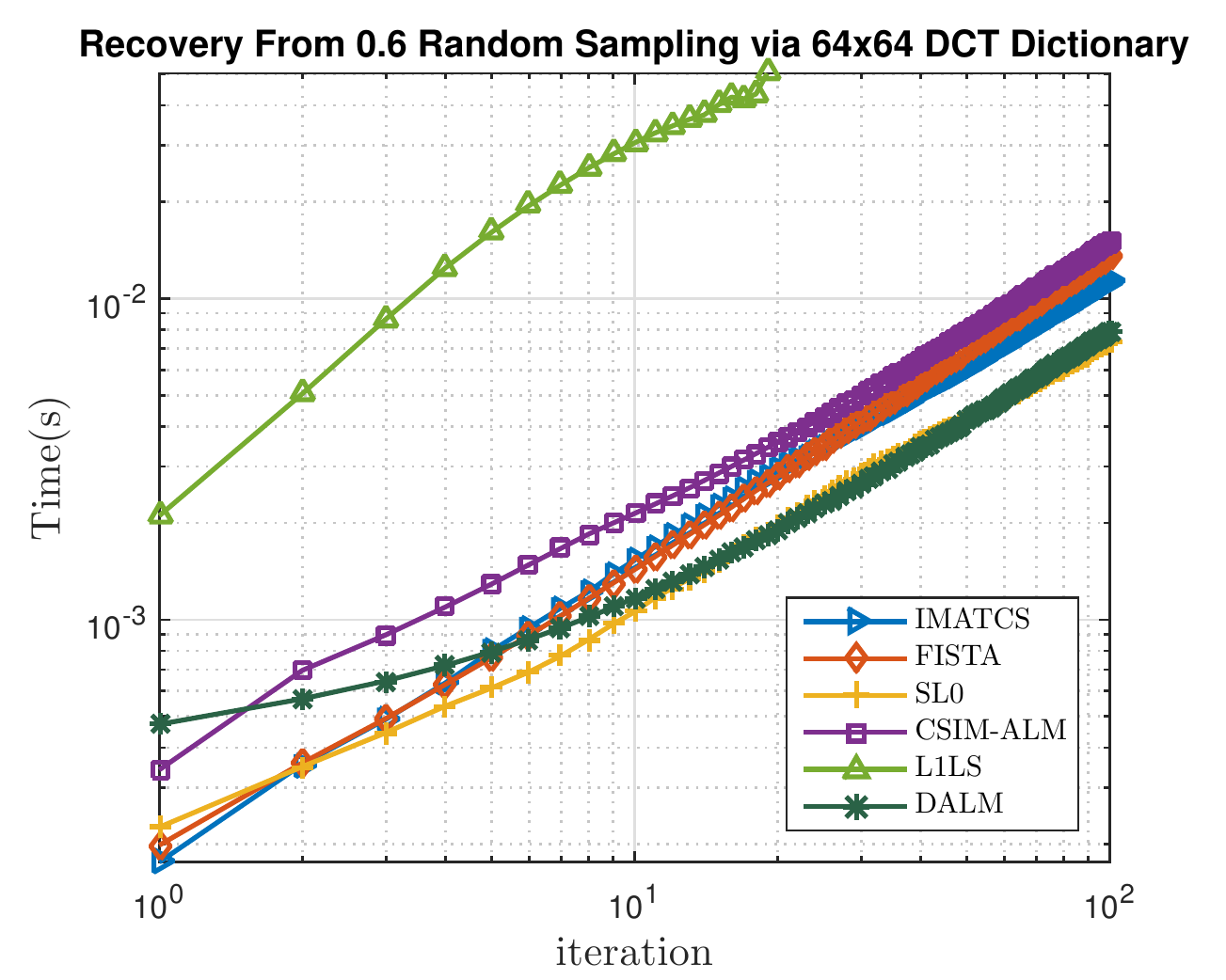}
\label{fig_time:2}
%\subfloat[$sr=0.4$]{\includegraphics*[width=2.2in]{Time_64_sr_0_4-eps-converted-to.pdf}\label{fig_time:3}}
\caption{Running time versus the number of iterations for sparse approximation via $64\times 64$ DCT atoms ($sr=0.6$).}
\label{fig_time}
\end{figure}

In this scenario, the patch signals are artificially produced by the product of generated sparse vectors $\{\svec_i\}$ with $(64\times 64)$ or $(64\times 128)$ DCT atoms as sparsifying basis. Similar to the previous experiment, we assume $10\%$ sparsity, i.e., $k=\lceil 0.1 p\rceil$ where $p=64$ or $128$ denotes the number of atoms. The $k$ nonzero coefficients of sparse vectors, are selected uniformly at random and the values are chosen according to Normal distribution.
For performance comparison, we consider relative error defined as:
\begin{equation}
\label{rel_err}
\mathrm{RelErr}_i = \dfrac{\norm \hat{\svec}_{i} - \svec_i \norm}{\norm \svec_i \norm}
\end{equation}
where $ \hat{\svec}_{i}$ denotes the recovered sparse vector, given the observation samples $\Hb \Db \svec
_{i}$ where $\Hb$ denotes the random sampling mask with sampling ratios $0.4, 0.6$ and $0.8$..
Fig. \ref{fig_err} shows these results for random sampling reconstruction of artificially generated signals versus the number of iterations.
As shown in this figure, the rate of convergence of CSIM-ALM is comparably faster than IMATCS and FISTA while the final error is significantly lower. This is specifically, more obvious for recovery via $(64\times64)$ DCT atoms. The running time of these methods for sparse reconstruction from complete DCT dictionary, is also given in Fig. \ref{fig_time}. The vertical axis in this figure represents the time (in seconds) consumed upon each iteration, and the horizontal axis shows the number of iterations.
Of course, the number of iterations for L1-LS which is determined by the stopping criterion of the algorithm, may be lass than the maximum limit. Hence, in Fig. \ref{fig_err} and Fig. \ref{fig_time} only some iterations of this method are given and beyond this limit the algorithm usually starts to diverge. According to Fig. \ref{fig_time}, the least complex algorithms are SL0, DALM and IMATCS but they need more iterations to reach acceptable recovery performance compared to CSIM-ALM. Indeed according to Fig. \ref{fig_err}, CSIM-ALM converges much faster than SL0 at high $SR$s and DALM at low sampling rates. This is in particular, more apparent for recovery via over-complete DCT dictionary where SL0 and L1-LS usually fail to converge.

\section{Conclusion}
\label{sec:conclusion}
In this paper, a fidelity metric called CSIM is introduced, which is convex and error-sensitive and can be applied for image quality assessment. The proposed index like MSE, is well suited for mathematical manipulations and like SSIM, has perceptual meanings. We investigate mathematical features of CSIM index in this paper and incorporate it as fidelity criterion for solving missing sample recovery problem based on sparse representation for the image signal. This recovery algorithm can also be applied in any sparse coding application such as dictionary learning and sparse approximation for image signals. Furthermore, an iterative ADMM-based algorithm is proposed to solve the \lone-minimization inverse problem for recovery of missing samples. The convexity of the optimization function leads to proof of convergence for the algorithm. Simulation results show the efficiency of CSIM and the the proposed iterative algorithm over counterpart methods for missing sample recovery of image signals.

\begin{appendices}
\section{Proof of Condition Number (Theorem \ref{theo:kappa})}
\label{section:App_kappa}
\subsubsection{Condition number of $\Wb$}
Let us assume $\Pb=\left(\frac{k_1}{n^2}-\frac{k_2}{n(n-1)}\right) \onevec_n \onevec_n^T$, $\mathbf{Q}=\Wb$ and $\gamma=\frac{k_2}{n-1}$. Now, using lemma \ref{lemm:1}, we can obtain the eigenvalues of $\Wb$. But first, we need to compute the eigenvalues of $\Pb$. Since $\Pb$ is symmetric and rank-one, it is implied that there is only a single (real) non-zero eigenvalue denoted by $\lambda_{\Pb_n}$ (with the corresponding eigenvector $\vvec_{\Pb_n}$) which satisfies:
\begin{equation}
\label{eq_10}
\Pb\vvec_{\Pb_n} =\left( \frac{k_1}{n^2}-\dfrac{k_2}{n(n-1)}\right)  \onevec_n \onevec_n^T \vvec_{\Pb_n}=\lambda_{\Pb_n} \vvec_{\Pb_n}
\end{equation}
It also implies that all the remaining (sorted) eigenvalues $\lambda_{\Pb_i},\, 1\leq i\leq n-1$ are zero.
Now, if we set $\vvec_{\Pb_n}=\onevec_n$, then $\lambda_{\Pb_n}$ is given by:
\begin{equation}
\label{eq_11}
\lambda_{\Pb_n} =\left( \frac{k_1}{n^2}-\frac{k_2}{n(n-1)}\right)  \onevec_n^T \onevec_n = \frac{k_1}{n}-\dfrac{k_2}{n-1}
\end{equation}
%Hence:
%\begin{equation}
%\label{eq_12}
%\lambda_{\Pb_i}=\left\{\!\!\begin{array}{ll}
%0,& i<n\\
%\frac{k_1}{n}-\frac{k_2}{n-1},& i=n
%\end{array}\right.
%\end{equation}
Consequently, the eigenvalues of $\mathbf{Q}$ are obtained by:
\begin{equation}
\label{eq_13}
\lambda_{\Qb_i}=\frac{k_2}{n-1}+\left\{\!\!\begin{array}{ll}
0,& i<n\\
\frac{k_1}{n}-\frac{k_2}{n-1},& i=n
\end{array}\right.
\end{equation}
Now assuming $k_2>k_1$, it yields $\kappa(\Wb) =\frac{\lambda_{\max}(\Wb)}{\lambda_{\min}(\Wb)}=\frac{k_2}{k_1}\left( \frac{n}{n-1}\right) = \frac{\theta_1}{n\theta_2+\theta_1}$.
\subsubsection{Condition number of $\tilde{\Db}$}
Assume $\Ab=\Db^T\Wb\Db\in \Real^{p\times p}$. Let $\Ab=\Bb+\Cb$ where $\Bb=\theta_1 \Db^T\Db$ and $\Cb =\theta_2 \Db^T \onevec \onevec^T \Db$. Also assume $\Db$ is full column rank ($p<n$) and the eigenvalues are sorted in descending order.
%\begin{enumerate}
Using Weyl's inequality \cite{Weyl} for the eigenvalues of sum of Hermitian matrices $\Bb$ and $\Cb$, we may write:
\begin{align}
\label{kappa_A}
\lambda_{\min}(\Ab) &\leq \lambda_{\min}\left(\Bb\right)+ \lambda_\Cb = \theta_1 \sigma^2_{\min}(\Db)+\lambda_\Cb \nonumber \\
\lambda_{\max}(\Ab) &\geq \lambda_{\max}\left(\Bb\right) = \theta_1 \sigma^2_{\max}(\Db)
\end{align}
where $\lambda_\Cb=\theta_2 \onevec^T \Db \Db^T \onevec$ is the only nonzero eigenvalue of $\Cb$.
Now using lemma \ref{lemm:kappa}, we have:
\begin{align}
\label{eq_kappa_mult}
\kappa(\tilde{\Db}) &= \kappa \left( (\Db^T\Wb\Db)^{-1} \Db^T \Wb \right) \\
& \leq \kappa \left( (\Db^T\Wb\Db)^{-1} \right) \kappa \left(\Db^T \right) \kappa \left(\Wb \right) \nonumber \\
&\leq \dfrac{\kappa \left(\Db\right) \kappa \left(\Wb \right)}{\kappa \left( \Db^T\Wb\Db\right)} \nonumber
\end{align}
Taking advantage of \eqref{kappa_A}, it is concluded that:
\begin{align}
\kappa(\tilde{\Db})&\leq \dfrac{\kappa(\Db)\kappa(\Wb)}{\kappa(\Ab)} \\
&\leq \,\kappa(\Db)\frac{k_2}{k_1}\left( \frac{n}{n-1}\right)\dfrac{\theta_1 \sigma^2_{\min}(\Db)+\theta_2 \onevec^T \Db \Db^T \onevec}{\theta_1 \sigma^2_{\max}(\Db)} \nonumber \\
&=\, \kappa(\Db)\frac{k_2}{k_1}\left( \frac{n}{n-1}\right) \Bigg(\!\!\left(\frac{k_1}{k_2}\left( \frac{n-1}{n^2}\right)-\frac{1}{n}\right) \frac{\onevec^T \Db \Db^T \onevec}{ \sigma^2_{\max}(\Db)} \nonumber \\
&\qquad\qquad\qquad\qquad\qquad\qquad\qquad\qquad\qquad +\frac{1}{\kappa^2(\Db)}\Bigg)\nonumber \\
&=\, \kappa(\Db)\frac{k_2}{k_1}\left( \frac{n}{n-1}\right) \left(\frac{1}{\kappa^2(\Db)}-\frac{1}{n} \frac{\onevec^T \Db \Db^T \onevec}{ \sigma^2_{\max}(\Db)}\right)\nonumber \\
&\qquad\qquad\qquad\qquad\qquad\qquad\,\,\, +\kappa(\Db)\left(\frac{1}{n} \frac{\onevec^T \Db \Db^T \onevec}{ \sigma^2_{\max}(\Db)}\right)\nonumber
\end{align}

\section{Proof of RIP Condition (Theorem \ref{theo:3})}
\label{section:App_RIP}
Assume $\Db'=\Wb^{1/2}\Db\in \Real^{n\times p}$ satisfies the RIP condition with constant $\delta'_{2k}$. If we denote the set of all $2k$-sparse vectors by $\Lambda_{2k}$ ($2k\leq p$), we have:
\begin{equation}
\forall \svec \neq\zerovec,  \svec\in \Lambda_{2k}, \quad(1-\delta'_{2k}) \leq \dfrac{\norm \Db' \svec\norm^2}{\norm \svec\norm^2} =\dfrac{\svec^T \Bb \svec}{\svec^T \svec}  \leq (1+\delta'_{2k})
\end{equation}
%For simplicity let $\Ab= \Db^T \Db$. Hence, for $\xvec\neq \zerovec$, the condition above may be restated as:
%\begin{equation}
%\forall \xvec \neq\zerovec,  \xvec\in \Lambda_{2k}, \quad(1-\delta_{2k}) \leq \dfrac{\xvec^T \Ab\xvec}{\xvec^T \xvec}  \leq (1+\delta_{2k})
%\end{equation}
%Now for $\Db'=\Wb^{1/2}\Db$ to satisfy RIP condition, there must be a constant $\delta'_{2k}$ such that:
%\begin{equation}
%\forall \xvec \neq\zerovec,  \xvec\in \Lambda_{2k}, \quad(1-\delta'_{2k}) \leq \dfrac{\xvec^T \Bb \xvec}{\xvec^T \xvec}  \leq (1+\delta'_{2k})
%\end{equation}
where $\Bb=\Db'^T \Db'=\Db^T \Wb\Db=\theta_1 \Db^T\Db +\theta_2 \Db^T \onevec \onevec^T \Db$. Also let $I_{2k}$ denote the set of indices corresponding with the non-zero elements of $\svec$ and $\Bb_{2k}=\Db_{2k}'^T \Db'_{2k}$ where $\Db'_{2k}=[\dvec'_i], i\in I_{2k}$ and $\dvec'_i$ denotes the $i$-th column of $\Db'$. Hence, incorporating the Min-Max theorem, it is sufficient to find $\delta'_{2k}$ which satisfies:
\begin{equation}
\max\{1- \lambda_{\min}(\Bb_{2k}),   \lambda_{\max}(\Bb_{2k})-1\} \leq \delta'_{2k}<1
\end{equation}
Now, similar to \cite{GCT}, using Gershgorin Circle Theorem (GCT) while assuming $\norm \dvec_i\norm=1,\, \forall i$, we may write:
\begin{align}
\label{CGT_B}
\vert \lambda_l(\Bb_{2k})-b_{i_{l},i_{l}}\vert \leq R_{i_{l}}(\Bb)=\!\!\!\!\!\sum_{j \in I_{2k}, j\neq i_{l}} &\!\!\!\!\!\!\vert b_{i_{l},j} \vert, \\
& i_{l} \in I_{2k},\, l=1,\ldots, 2k\nonumber
\end{align}
%where the elements $b_{i,j}$ of $\Bb$ are obtained by $b_{i,j}=\theta_1 \dvec_i^T \dvec_j+ \theta_2 \dvec_i^T \onevec \onevec^T \dvec_j$.
To avoid $\lambda_{\min}(\Bb_{2k})=0$, we assume $\Bb_{2k}$ is full rank, i.e., $n\geq 2k$. Thus, using \eqref{CGT_B}, the upper bound for the maximum eigenvalue of $\Bb_{2k}$ is obtained as follows:
\begin{align}
\lambda_{\max}(\Bb_{2k}) &\leq \max_{i\in I_{2k}} \{ b_{i,i}+ R_i(\Bb)\} \nonumber \\
&= \max_{i\in I_{2k}} \Big\{ \theta_1 \norm \dvec_i \norm^2+ \theta_2 (\dvec_i^T \onevec)^2+\!\!\!\!\sum_{j \in I_{2k}, j\neq i} \!\!\!\!\!\!\vert \theta_1 \dvec_i^T \dvec_j\nonumber \\
&\qquad\qquad\qquad\qquad\qquad\qquad\quad+ \theta_2 \dvec_i^T \onevec \onevec^T \dvec_j \vert \Big\} \nonumber \\
& \leq \max_{i\in I_{2k}} \Big\{\theta_1 + \theta_2 (\norm \dvec_i\norm\norm \onevec\norm)^2\nonumber \\
&\quad\quad\quad\quad+ \!\!\!\!\!\!\sum_{j \in I_{2k}, j\neq i} \!\!\!\!\!\!\left(\vert \theta_1\vert \vert \dvec_i^T \dvec_j \vert+\vert \theta_2\vert \vert \dvec_i^T \onevec \onevec^T \dvec_j \vert\right) \!\!\Big\}\nonumber \\
&\leq \theta_1 + n\theta_2 +   (2k-1) \max_{j\neq i} \Big\{\vert \theta_1\vert  \vert\dvec_i^T \dvec_j \vert\nonumber \\
&\qquad\qquad\qquad\qquad\qquad\,\qquad+ \vert \theta_2\vert \vert \dvec_i^T \onevec \onevec^T \dvec_j \vert \Big\} \nonumber \\
&\leq \theta_1 + n\theta_2 + (2k-1)\vert \theta_1\vert  \mu (\Db)+(2k-1)\times\nonumber \\
& \quad\Big( \vert \theta_2 \vert \max_{j\neq i} \{ \vert \dvec_i^T \dvec_j \vert + \big| \sum_{p}\sum_{ q\neq p}  d_{i,p}d_{j,q} \big|\} \Big)\nonumber \\
&\leq \theta_1 + n\theta_2 +    (2k-1)\times\nonumber \\
&\quad \Big( \vert \theta_1\vert \mu (\Db)+ \vert \theta_2 \vert (\mu(\Db)+(n-1)\norm\dvec_i\norm \norm \dvec_j \norm) \Big) \nonumber \\
&\leq \theta_1 + n\theta_2+(n-1)(2k-1) \vert \theta_2 \vert \nonumber\\
&\qquad\qquad \quad+(\vert \theta_1\vert +\vert \theta_2\vert) (2k-1) \mu (\Db)
\end{align}
where we have applied the Cauchy-Schwarz inequality for $\big| \sum_{p}\sum_{ q\neq p}  d_{i,p}d_{j,q} \big| \leq (n-1)\norm\dvec_i\norm \norm \dvec_j \norm$. Furthermore, for the minimum eigenvalue, we can write:
\begin{align}
\lambda_{\min}(\Bb_{2k}) &\geq \min_{i\in I_{2k}}  \{b_{i,i}- R_i(\Bb) \}\nonumber \\
&= \min_{i\in I_{2k}} \Big\{\theta_1 \norm \dvec_i \norm^2+ \theta_2 (\dvec_i^T \onevec)^2\nonumber \\
&\qquad\qquad\qquad\,\,\,-\!\!\!\!\!\sum_{j \in I_{2k}, j\neq i}\!\!\!\!\!\! \vert \theta_1 \dvec_i^T \dvec_j+ \theta_2 \dvec_i^T \onevec \onevec^T \dvec_j \vert\Big\} \nonumber \\
& \geq  \theta_1 -\max_{i\in I_{2k}}  \!\!\!\!\sum_{j \in I_{2k}, j\neq i} \!\!\!\!\! \left(\vert \theta_1\vert \vert \dvec_i^T \dvec_j \vert+ \vert \theta_2\vert \vert \dvec_i^T \onevec \onevec^T \dvec_j \vert\right) \nonumber \\
&\geq \theta_1 -(n-1)(2k-1) \vert \theta_2 \vert \nonumber \\
&\qquad\qquad\qquad-  (\vert \theta_1\vert +\vert \theta_2\vert) (2k-1) \mu (\Db)
\end{align}
Thus, it is sufficient to choose $ \delta'_{2k}$ such that:
\begin{align}
\max\{1- \theta_1 ,  \theta_1 &+ n\theta_2-1 \}+(n-1)(2k-1) \vert \theta_2 \vert \nonumber \\
&+ (\vert \theta_1\vert +\vert \theta_2\vert) (2k-1) \mu (\Db) \leq \delta'_{2k}<1
\end{align}
Assume $k_2 = n-1$ and $\rho=k_2/k_1>1$. Substituting $\theta_1$ and $\theta_2$ from \eqref{eq_8}, we yield $\theta_1=1$ and $\theta_2<0$. Hence, we can simplify the condition above as follows:
\begin{align}
\!\!(n-1)(2k-1) \Big( \frac{1}{n}\!-\!\frac{n-1}{\rho n^2}\Big)+ (2k-1) \mu (\Db) \Big( \frac{n+1}{n}- &\frac{n-1}{\rho n^2}\Big) \nonumber \\
&\leq \delta'_{2k}
\end{align}
which leads to:
%Now satisfying $0\leq\delta'_{2k} <1$, we finally yield:
\begin{align}
1< \rho \leq \frac{C_1}{C_2-\delta'_{2k}}
\end{align}
where $C_1$ and $C_2$ are obtained by:
\begin{align}
C_1 &= \frac{1}{n^2}(2k-1)(n-1)(n-1+\mu(\Db))\nonumber \\
C_2 &= \frac{2k-1}{n}(n-1+\mu(\Db)(n+1))
\end{align}
If  $n>1$ and $k>\frac{1}{2}(1+\frac{n}{n-1})>1$, we have $C_1>0$ and $C_2>1>\delta'_{2k}$. Now to fulfil the assumption $\rho>1$, we must have $C_1>C_2-\delta'_{2k}$ which implies:

\begin{equation}
\label{eq_ineq_mu}
h(n)=n^2\left(\frac{\delta'_{2k}}{2k-1}-\mu(\Db)\right)-n+1-\mu(\Db)>0
\end{equation}
Since $h(n)$ is quadratic with respect to $n$, it has two roots specified by $n_{\min}$ and $n_{\max}$ where:
\begin{align}
n_{\min} = \frac{1}{2}\left(\frac{\delta'_{2k}}{2k-1}-\mu(\Db)\right)^{\!\!-1}\!\!\Big(1-\sqrt{\Delta(\mu(\Db))}\Big) \nonumber \\
n_{\max}= \frac{1}{2}\left(\frac{\delta'_{2k}}{2k-1}-\mu(\Db)\right)^{\!\!-1}\!\!\Big(1+\sqrt{\Delta(\mu(\Db))}\Big)
\end{align}
The function $\Delta(\mu(\Db))= 1-4\left(\frac{\delta'_{2k}}{2k-1}-\mu(\Db)\right)\big(1-\mu(\Db)\big)$ itself, has two roots. The greater root satisfies $\mu_{\max}>1$ and the smaller one obtained by:
\begin{align}
\mu_{\min}=\frac{1}{2} \Bigg(1+\frac{\delta'_{2k}}{2k-1}-\sqrt{1+\left(1-\frac{\delta'_{2k}}{2k-1}\right)^2}\Bigg)
\end{align}
satisfies $\mu_{\min}<\frac{\delta'_{2k}}{2k-1}$. Now, since $\frac{\delta'_{2k}}{2k-1}<1$ for $k>1$, we will have the following cases:

\begin{enumerate}

\item  If $0<\mu(\Db)<\max\{0,\mu_{\min}\}$: Then $\Delta(\mu(\Db))<0$ and for any $n\geq 2k$, $h(n)$ is positive. But for $\mu_{\min}$ to be positive, we need to have $k<\frac{4\delta'_{2k}+1}{2}$ which only occurs for $k=2$ and $\delta'_{2k}>\frac{3}{4}$.

\item $\max\{0,\mu_{\min}\} <\mu(\Db)<\frac{\delta'_{2k}}{2k-1}$: In this case $0<\Delta(\mu(\Db))<1$ and the function $h(n)$ is positive if
$n> n_{\max}$ or $n< n_{\min}$.

\item $\frac{\delta'_{2k}}{2k-1}< \mu(\Db)<1$: In this case $\Delta(\mu(\Db))>1$ and hence $h(n)$ has two distinct roots. But only $n_{\min}$ is positive and thus inequality \eqref{eq_ineq_mu} holds if $n< n_{\min}$.

\end{enumerate}

With our initial assumptions about the values of $k$ and $n$, the case $n<n_{\min}$ is not feasible. Hence, considering $\delta_{2k}<\sqrt{2}-1$, the only feasible case is when $0 <\mu(\Db)<\frac{\delta'_{2k}}{2k-1}$ and $n>n_{\max}$ where $k$ satisfies:

\begin{equation}
\label{eq_inq_k}
2k\leq \min \left\{ \left( 1+\frac{\delta'_{2k}}{\mu(\Db)}\right), \min \{n,p\} \right\}
\end{equation}

%There are various bibliography styles available. You can select the style of your choice in the preamble of this document. These styles are Elsevier styles based on standard styles like Harvard and Vancouver. Please use Bib\TeX\ to generate your bibliography and include DOIs whenever available.
%
%Here are two sample references: \cite{Feynman1963118,Dirac1953888}.
\section{Proof of Convergence (Theorem \ref{theo:4})}
\label{section:App_Proof}
We use similar procedure as in \cite{Yang11} for the proof of convergence of the proposed algorithm (Alg. \ref{Algorithm_0}) with fixed regularizing parameter $\alpha$. Before we proceed, let us define two notations used in this proof. We define $\delta \uvec^{(t+1)}= \uvec^{(t+1)}-\uvec^{(t)}$ and $\Delta \uvec^{(t+1)}= \uvec^{(t+1)}-\uvec^{\ast}$ where $\uvec^{\ast} = \left(\svec^{\ast}, \zvec^{\ast}, \mub^{\ast}_1, \mub^{\ast}_2\right)^T$ is the optimal solution to \eqref{eq_21}. Since CSIM is convex, the cost function is also convex. Hence, the KKT optimality conditions for problem \eqref{eq_21} yields its global optimum:
\begin{align}
\label{eq_KKT_21}
\mub^\ast_1 -\Hb^T \mub^\ast_2 &=\zerovec \nonumber \\
\frac{1}{\alpha}\Db^T \mub^\ast_1   &\in \partial \norm \svec^{\ast} \norm_1 \nonumber \\
2(\Wb+ \gamma \Ib)  \zvec^\ast &= -\mub^\ast_2 \nonumber \\
\xvec^\ast &= \Db \svec^\ast \nonumber \\
\zvec^\ast &= \Hb \xvec^\ast-\yvec
\end{align}
Consider equation \eqref{eq_23} for the update of $\xvec$. Substituting $\mub^{(t)}_1$ and $\mub^{(t)}_2$ from \eqref{eq_40} and incorporating $\mub_1^\ast -\Hb^T \mub_2^\ast =\zerovec$, we get:
%\begin{equation}
%\label{eq_81}
%-\sigma_1 \Db \delta \svec^{(t+1)}-\sigma_2 \Hb^T \delta \zvec^{(t+1)} -\mub_1^{(t+1)}+\Hb^T \mub_2 ^{(t+1)} = \zerovec
%\end{equation}
\begin{equation}
\label{eq_xup}
\sigma_1 \Db \delta \svec^{(t+1)}+\sigma_2 \Hb^T \delta \zvec^{(t+1)} +	\Delta \mub_1^{(t+1)}-\Hb^T  \Delta \mub_2^{(t+1)}  = \zerovec
\end{equation}
Furthermore, optimizing problem \eqref{eq_28} with respect to $\svec$, while $\svec_0 = \svec^{(t)}$, we obtain the following:
\begin{equation}
\label{eq_subg}
\frac{1}{\alpha}\left( \Db^T \mub^{(t+1)}_1 +\sigma_1 (\Db^T \Db- \lambda \Ib) \delta  {\svec^{(t+1)}} \right) \in \partial \norm \svec^{(t+1)} \norm_1
\end{equation}
Now, according to lemma \ref{lemm:2}, we have:
\begin{equation}
 (\svec ^{(t+1)}-\svec^{\ast})^T\left(\partial \norm \svec ^{(t+1)} \norm_1 - \partial \norm \svec^{\ast} \norm_1\right) \geq 0
\end{equation}
Using $ \partial \norm \svec^{\ast} \norm_1$ from \eqref{eq_KKT_21} and $\partial \norm \svec ^{(t+1)} \norm_1$ from \eqref{eq_subg} and discarding $\alpha>0$, we can conclude that:
\begin{equation}
\Delta {\svec^{(t+1)}} ^T \left[ \Db^T  \Delta {\mub^{(t+1)}_1} +\sigma_1 (\Db^T \Db- \lambda \Ib) \delta  {\svec^{(t+1)}}   \right]  \geq 0
\end{equation}
Next, substituting $\Delta \mub_1^{(t+1)}$ from \eqref{eq_xup} we have:
\begin{align}
\Delta {\svec^{(t+1)}} ^T\!\Big\{ \Db^T \Big[\!\!- \sigma_1 \Db &\delta \svec^{(t+1)} - \sigma_2 \Hb^T \delta \zvec^{(t+1)}+\Hb^T  \Delta \mub_2^{(t+1)} \Big] \nonumber \\
&+\sigma_1 \Db^T \Db  \delta  {\svec^{(t+1)}}  - \lambda \sigma_1  \delta  {\svec^{(t+1)}}  \Big\} \geq 0
\end{align}
or equivalently:
\begin{align}
\label{eq_82}
- \left( \Hb \Db \Delta {\svec^{(t+1)}} \right) ^T \Big[\sigma_2  \delta \zvec^{(t+1)}&- \Delta \mub_2^{(t+1)} \Big]  \nonumber \\
&-\lambda \sigma_1 \Delta {\svec^{(t+1)}}^T \delta  {\svec^{(t+1)}}  \geq 0
\end{align}
Now, using equation \eqref{eq_38} for the update of $\zvec$ as well as the optimality condition $2(\Wb + \gamma \Ib) \zvec ^{\ast} =- \mub^{\ast}_2$, we yield:
\begin{equation}
\label{eq_zup}
2(\Wb+\gamma \Ib ) \Delta {\zvec^{(t+1)}} = -\Delta {\mub^{(t+1)}_2}
\end{equation}
On the other side, incorporating the Lagrange multipliers update equations from \eqref{eq_40}, we obtain:
\begin{align}
\Hb \Db \Delta {\svec^{(t+1)}} &= \Hb \Delta {\xvec^{(t+1)}} - \frac{1}{\sigma_1}  \Hb \delta {\mub^{(t+1)}_1} \nonumber \\
&= \Delta {\zvec^{(t+1)}} - \frac{1}{\sigma_2}  \delta {\mub^{(t+1)}_2}- \frac{1}{\sigma_1}  \Hb \delta {\mub^{(t+1)}_1}
\end{align}
Hence, expanding \eqref{eq_82}, we have:
\begin{align}
\label{eq_inq_1}
\delta {\mub^{(t+1)}_2}^T \!\!\delta &{\zvec^{(t+1)}}  -\sigma_2 \!\Delta {\zvec^{(t+1)}}^T \delta {\zvec^{(t+1)}} -\frac{1}{\sigma_2}  \delta {\mub^{(t+1)}_2}^T \!\!\Delta {\mub^{(t+1)}_2} \nonumber \\
&+\frac{1}{\sigma_1} \delta {\mub^{(t+1)}_1}^T  \left[-\Hb^T \Delta {\mub^{(t+1)}_2}+ \sigma_2 \Hb^T \delta {\mub^{(t+1)}_2}\right]\nonumber \\
&+ \Delta {\zvec^{(t+1)}}^T  \Delta {\mub^{(t+1)}_2}   -\lambda \sigma_1 \Delta {\svec^{(t+1)}}^T \delta {\svec^{(t+1)}}
 \geq 0
\end{align}
Now, we define:
\begin{equation}
\Gb_0 \!=\!   \left(\! \begin{array}{cccc}
\!\!\lambda \sigma_1 \Ib &\!\!\! &\!\!\! &\!\!\!\\
\!\! &\!\!\! \sigma_2 \Ib  &\!\!\!  &\!\! \\
\!\! &\!\!\! & \!\!\!\frac{1}{\sigma_1}\Ib &\!\! \\
\!\! &\!\!\! &\!\!\! & \!\!\!\frac{1}{\sigma_2} \Ib
\end{array} \!\!\right) \,\, \Gb \!=\!  \left( \! \begin{array}{cccc}
\!\!\lambda \sigma_1 \Ib &\!\!\! &\!\!\! &\!\!\\
\!\! & \!\!\!\zeta \sigma_2 \Ib &\!\!\! &\!\!\\
\!\! &\!\!\! &\!\!\! \frac{1}{\sigma_1}\Ib &\!\! \\
\!\! &\!\!\! &\!\!\! &\!\!\! \frac{1}{\sigma_2} \Ib
\end{array} \!\!\right)
\end{equation}
where $\zeta>1$.
Substituting the brackets in \eqref{eq_inq_1} from \eqref{eq_xup}, while incorporating \eqref{eq_zup}, we finally yield:
\begin{align}
\label{eq_inqall}
-\Delta {\uvec^{(t+1)}}\Gb_0 \delta {\uvec^{(t+1)}}\geq & -\delta {\mub^{(t+1)}_2} ^T\!\delta {\zvec^{(t+1)}} \!+ \delta {\mub^{(t+1)}_1} ^T\! \Db \delta {\svec^{(t+1)}} \nonumber \\
&+2 \Delta {\zvec^{(t+1)}}^T(\Wb+\gamma \Ib )\Delta {\zvec^{(t+1)}}
\end{align}
Multiplying \eqref{eq_inqall} by $2$, adding the term $-(\zeta-1)\sigma_2 \Delta {\zvec^{(t+1)}}^T\delta {\zvec^{(t+1)}}^T$ to both sides of the inequality and finally using the Cauchy-Schwartz inequality $2\avec^T \bvec \geq -\frac{1}{\beta} \norm \avec \norm^2 - {\beta} \norm \bvec \norm^2$ ($\beta>0$), we get the following:
\begin{align}
\label{eq_92}
-2\Delta {\uvec^{(t+1)}}\Gb \delta {\uvec^{(t+1)}}\geq & - \left[ \frac{1}{\beta_1} \norm \delta {\mub^{(t+1)}_1}\norm ^2 +\beta_1 \norm \Db \delta {\svec^{(t+1)}} \norm^2 \right] \nonumber \\
&- \left[ \frac{1}{\beta_2} \norm \delta {\mub^{(t+1)}_2}\norm ^2 +\beta_2 \norm \delta {\zvec^{(t+1)}} \norm^2 \right] \nonumber \\
&-(\zeta-1)\sigma_2 \Big[ \frac{1}{\beta_3} \norm \delta {\zvec^{(t+1)}}\norm ^2 \nonumber \\
&\qquad\qquad\qquad\qquad+\beta_3 \norm \Delta {\zvec^{(t+1)}} \norm^2 \Big] \nonumber \\
&+4 \Delta {\zvec^{(t+1)}}^T(\Wb+\gamma \Ib )\Delta {\zvec^{(t+1)}}
\end{align}
If $\beta_3$ and $\zeta$ are chosen such that $4(\Wb+\gamma \Ib )-\beta_3(\zeta-1)\sigma_2 \Ib$ is positive semi-definite, then we can discard the terms involving $\Delta {\zvec^{(t+1)}}$, in right hand side of \eqref{eq_92}. To fulfil this, we must have $4\lambda_{\min}(\Wb) + 4\gamma - \beta_3 (\zeta-1)\sigma_2\geq0$.
Now, using \eqref{eq_13} with $k_2 = n-1$ and $\rho=k_2/k_1>1$, we conclude that:
\begin{equation}
\label{eq_in_rho1}
\beta_3\leq\frac{4}{(\zeta-1)\sigma_2}\left(\gamma+\frac{n-1}{\rho n}\right)
\end{equation}
Next, using $\Delta {\uvec^{(t)}} =\Delta {\uvec^{(t+1)}} -\delta {\uvec^{(t+1)}}$  we proceed as follows:
\begin{align}
\label{eq_inq_eta}
 \norm \Delta {\uvec^{(t)}} \norm_{\Gb}^2 \!-\!  \norm \Delta {\uvec^{(t+1)}} \norm_{\Gb}^2 &=-2\Delta {\uvec^{(t)}}\Gb \delta {\uvec^{(t+1)}} \!-\! \norm \delta {\uvec^{(t+1)}} \norm_{\Gb}^2 \nonumber \\
 &=-2\Delta {\uvec^{(t+1)}}\Gb \delta {\uvec^{(t+1)}}\nonumber \\
 &\quad\,+2 \norm \delta {\uvec^{(t+1)}} \norm_{\Gb}^2- \norm \delta {\uvec^{(t+1)}} \norm_{\Gb}^2 \nonumber \\
 &\geq \eta  \norm \delta {\uvec^{(t+1)}} \norm_{\Gb}^2
\end{align}
where the constant $\eta>0$ determines the rate of convergence of the algorithm. Combining \eqref{eq_92} and \eqref{eq_inq_eta} and discarding the terms with $\Delta {\zvec^{(t+1)}}$, we have:
\begin{align}
\label{eq_eta}
\lambda \sigma_1 - \beta_1 \norm \Db \norm^2 &\geq\eta \lambda \sigma_1  \nonumber \\
\zeta \sigma_2 - \frac{1}{\beta_3} (\zeta-1) \sigma_2 - \beta_2 &\geq \eta \zeta \sigma_2 \nonumber  \\
\frac{1}{\sigma_1}-\frac{1}{\beta_1}&\geq\eta \frac{1}{\sigma_1} \nonumber\\
\frac{1}{\sigma_2}-\frac{1}{\beta_2}&\geq\eta \frac{1}{\sigma_2}
\end{align}
From $\eta>0$, we conclude that $\sigma_1 <\beta_1 <\frac{\lambda}{\norm \Db \norm^2}\sigma_1$ and $\sigma_2 <\beta_2<\sigma_2 (\zeta - \frac{1}{\beta_3} (\zeta-1))$; which in turn implies $\lambda> \lambda_{\max} (\Db)=\norm\Db \norm^2 $ and
$1<\beta_3$.
Now, define $\delta_1 = \frac{\lambda}{\norm \Db\norm^2} -1$, $\beta_1 = \sigma_1 (1+r_1\delta_1)$, $\delta_2 = (\zeta-1)(1-1/\beta_3)$, $\beta_2 = \sigma_2(1+r_2\delta_2)$, $\delta_3= \frac{\sigma_2}{4(\gamma+\frac{n-1}{\rho n})}$ and $\zeta = 1+\frac{1}{r_3 \delta_3}$.
Subsisting these into \eqref{eq_eta}, taking advantage of \eqref{eq_in_rho1}, it implies that we if choose $0<r_1, r_2<1$ and $r_3>1$, then there will exist $\eta>0$ which satisfies \eqref{eq_inq_eta} and is obtained by:
\begin{align}
0<\eta \leq  \min &\Bigg\{\dfrac{r_1 \delta_1}{1+r_1 \delta_1}, \dfrac{(1-r_1) \delta_1}{1+\delta_1},\dfrac{r_2 \delta_2}{1+r_2 \delta_2}, \\
&\left(1+\frac{(1-r_2)(r_3-1)}{r_3^2 \delta_3}\right)\frac{r_3 \delta_3}{1+r_3 \delta_3}\Bigg\} \nonumber
\end{align}
Thus, using \eqref{eq_inq_eta} to prove the convergence, we may write:
\begin{align}
\eta \sum_{t=0}^{\infty} \norm \delta {\uvec^{(t+1)}} \norm_{\Gb}^2 &\leq \sum_{t=0}^{\infty} \left\{ \norm \Delta {\uvec^{(t)}} \norm_{\Gb}^2 -  \norm \Delta {\uvec^{(t+1)}} \norm_{\Gb}^2 \right\}  \\
&=\norm \Delta {\uvec^{(0)}} \norm_{\Gb}^2 -  \norm \Delta {\uvec^{(\infty)}} \norm_{\Gb}^2 \leq  \norm \Delta {\uvec^{(0)}} \norm_{\Gb}^2\nonumber
\end{align}
Since $\sum_{t=0}^{\infty} \norm \delta {\uvec^{(t+1)}} \norm_{\Gb}^2>0$ is bounded from above, we can conclude  that $ \norm \delta {\uvec^{(t)}} \norm_{\Gb} \rightarrow 0$ or equivalently $\delta \uvec^{(t)} \rightarrow \zerovec$ (since $\Gb$ is positive definite). This implies that the algorithm converges to a solution denoted by ${\tilde{\uvec}} = \lim_{t \rightarrow\infty}\uvec^{(t)}$. Now, since $\delta \mub^{(t)}_1\rightarrow \zerovec$ and $\delta \mub^{(t)}_2\rightarrow \zerovec$,  from \eqref{eq_40} we obtain:
\begin{align}
\label{eq_80}
 \sigma_1 \left(\xvec^{(t+1)} - \Db \svec^{(t+1)}\right) &=\delta \mub^{(t)}_1 \rightarrow \zerovec \nonumber \\
\sigma_2 \left(\zvec^{(t+1)} - \Hb \xvec^{(t+1) }+ \yvec\right) &= \delta \mub^{(t)}_2 \rightarrow \zerovec
\end{align}
which implies $\tilde{\xvec}= \Db \tilde{\svec}$ and $\tilde{\zvec}= \Hb \tilde{\xvec}-\yvec$. Also using \eqref{eq_38} and \eqref{eq_xup} we get:
\begin{align}
 2(\Wb+ \gamma \Ib)  \zvec^{(t+1)}& =  -\mub_2^{(t+1)} \nonumber \\
\mub_1^{(t+1)}-\Hb^T \mub_2 ^{(t+1)} &=-\sigma_1 \Db \delta \svec^{(t+1)}-\sigma_2 \Hb^T \delta \zvec^{(t+1)}  \rightarrow \zerovec
\end{align}
Hence, we can conclude that $\tilde{\mub}_1 = -\Hb^T \tilde{\mub}_2$ and $2(\Wb+ \gamma \Ib)  \tilde{\zvec} = -\tilde{\mub}_2$. Together with \eqref{eq_80} this implies that every limit point of $\{\uvec^{(t)} \}$ is an optimal solution to \eqref{eq_21}, i.e., $\uvec^{(t)} \rightarrow \uvec^\ast$.

It can also be easily verified that any solution to \eqref{eq_21} is also a solution to \eqref{eq_20_equi}.
\section{Lemmas and Proofs}
\begin{lemma}
\label{lemm:1}
Let $\mathbf{Q}=\Pb+\gamma \Ib_n$ where $\Pb,\, \Qb \in \Real^{n\times n}$. If all the eigenvalues of $\Pb$ are given as $\lambda_{\Pb_1}, \lambda_{\Pb_2}, \ldots  \lambda_{\Pb_n}$, then the eigenvalues of $\mathbf{Q}$ will be obtained using the following equation:
\begin{equation}
\label{eq_9}
\lambda_{\mathbf{Q}_i}=\gamma + \lambda_{\Pb_i}
\end{equation}
\end{lemma}
\begin{IEEEproof}
Let $\vvec_{\Pb_i}$ be an eigenvector of $\Pb$. We thus have:
\begin{equation}
\Qb \vvec_{\Pb_i} = \Pb \vvec_{\Pb_i}+\gamma \vvec_{\Pb_i}= (\lambda_{\Pb_i} + \gamma) \vvec_{\Pb_i}
\end{equation}
which implies that $\vvec_{\Qb_i}=\vvec_{\Pb_i}$ is an eigenvector and $\lambda_{\mathbf{Q}_i} = \lambda_{\Pb_i} + \gamma$ is an eigenvalue of $\Qb$.
\end{IEEEproof}
\begin{lemma}
\label{lemm:kappa}
For any matrices $\Ab$ and $\Bb$ with $\kappa$ defined in \eqref{eq_kappa_def}, we have:
\begin{equation}
\label{eq_9}
\kappa(\Ab \Bb) \leq \kappa (\Ab) \kappa (\Bb)
\end{equation}
\end{lemma}
\begin{IEEEproof}
Using the the definitions of $\sigma_{\min} (\Ab)= \min_{\xvec \neq \zerovec} \frac{\norm \Ab \xvec \norm}{\norm \xvec \norm}$ and $\sigma_{\max} (\Ab)= \max_{\xvec \neq \zerovec} \frac{\norm \Ab \xvec \norm}{\norm \xvec \norm}$ it is implied that $\sigma_{\min} (\Ab) \norm \xvec \norm\leq \norm \Ab \xvec \norm \leq \sigma_{\max} (\Ab) \norm \xvec\norm$. Consequently if we substitute $\xvec$ by $\Bb\yvec$, it is easy to show that $\sigma_{\min} (\Ab) \sigma_{\min} (\Bb) \norm \xvec \norm\leq \norm \Ab \Bb \xvec \norm \leq \sigma_{\max} (\Ab) \sigma_{\max} (\Bb)\norm \xvec\norm$ which concludes the proof.
\end{IEEEproof}
\begin{lemma}
\label{lemm:2}
For any two vectors $\svec^{\ast}$ and $\svec ^{(t+1)}$, we have:
\begin{equation}
 (\svec ^{(t+1)}-\svec^{\ast})^T\left(\partial \norm \svec ^{(t+1)} \norm_1 - \partial \norm \svec^{\ast} \norm_1\right) \geq 0
\end{equation}
\end{lemma}
\begin{IEEEproof}
Since \lone~norm is convex, exploiting the definition of sub-gradient of $f=\norm \svec \norm_1$ at point $\svec^{\ast}$, we obtain:
\begin{equation}
\label{eq_lemma_2_1}
\norm \svec ^{(t+1)} \norm_1 \geq \norm \svec^{\ast} \norm_1 +\left( \svec ^{(t+1)} -\svec^{\ast} \right)^T \partial \norm \svec^{\ast} \norm_1
\end{equation}
Next, choose $\svec ^{(t+1)}$ as the starting point and $\svec ^{\ast}$ as the test point. Using a similar inequality, yields:
\begin{equation}
\label{eq_lemma_2}
\norm \svec^{\ast} \norm_1 \geq \norm  \svec ^{(t+1)}  \norm_1 +\left( \svec^{\ast} -\svec ^{(t+1)}  \right)^T \partial \norm\svec ^{(t+1)} \norm_1
\end{equation}
Now, reversing \eqref{eq_lemma_2_1} and adding it to \eqref{eq_lemma_2}, we get:
\begin{equation}
 (\svec ^{(t+1)}-\svec^{\ast})^T\left(\partial \norm \svec ^{(t+1)} \norm_1 - \partial \norm \svec^{\ast} \norm_1\right) \geq 0
\end{equation}
\end{IEEEproof}
\end{appendices}

%\bibliography{mybibfile}

\end{document}